\documentclass[10pt,twocolumn,letterpaper]{article}

\usepackage{iccv}
\usepackage{times}
\usepackage{epsfig}
\usepackage{graphicx}
\usepackage{amsmath}
\usepackage{amssymb}
\usepackage{microtype}
 
\usepackage[ruled, linesnumbered]{algorithm2e}
\usepackage{booktabs}       
\usepackage{enumitem}
\usepackage{multirow}
\usepackage[dvipsnames]{xcolor}
\usepackage{subcaption}
\usepackage{wrapfig}
\usepackage{scalerel}
\usepackage{nicefrac}
\usepackage{bbm}
\usepackage{textcomp}
\graphicspath{{figures/}}
\setlist[itemize]{itemsep=0pt,topsep=0pt,partopsep=0pt,parsep=0pt,leftmargin=4mm}

\usepackage[pagebackref=true,breaklinks=true,colorlinks,
bookmarks=false]{hyperref}

\newcommand{\red}[1]{\textcolor{red}{#1}}

\definecolor{mygreen}{RGB}{0, 175, 0}
\definecolor{myblue}{RGB}{0, 102, 204}
\definecolor{mygray}{RGB}{64, 64, 64}
\newcommand{\myblue}[1]{\textcolor{myblue}{#1}}
\newcommand{\mygreen}[1]{\textcolor{mygreen}{#1}}
\SetKwInput{KwInit}{Initialization}

  \def\RR{{\mathbb R}}

    \def\cT{{\mathcal T}} 
\def\cC{{\mathcal C}}

  \def\cL{{\mathcal L}}

\newcommand{\mask}{\mathbf{s}^\alpha}
\newcommand{\omask}{\mathbf{o}^\alpha}
\newcommand{\proto}{\mathbf{s}}
\newcommand{\papp}{\mathbf{s}^c}

\newcommand{\bkg}{\mathbf{b}}
\newcommand{\obj}{\mathbf{o}}

\newcommand{\oapp}{\mathbf{o}^c}
\newcommand{\inp}{\mathbf{x}}
\newcommand{\comp}{\mathbf{c}}

\newcommand{\occ}{\delta}

\newcommand{\dpg}{\phi}
\newcommand{\dpp}{\psi}

\DeclareMathOperator*{\softclip}{softclip}
\DeclareMathOperator*{\clip}{clip}
\newcommand{\tfp}{\cT^{\,\textrm{\scriptsize spr}}}
\newcommand{\tfl}{\cT^{\,\textrm{\scriptsize lay}}}
\newcommand{\tfb}{\cT^{\,\textrm{\scriptsize bkg}}}
\newcommand{\parp}{\nu}
\newcommand{\parl}{\eta}

\iccvfinalcopy 

\def\iccvPaperID{4323} 
\def\httilde{\mbox{\tt\raisebox{-.5ex}{\symbol{126}}}}

\begin{document}

\title{Unsupervised Layered Image Decomposition into Object Prototypes}

\author{Tom Monnier\textsuperscript{1} \quad Elliot Vincent\textsuperscript{1,2} \quad Jean 
  Ponce\textsuperscript{2,3} \quad Mathieu Aubry\textsuperscript{1}\\
\textsuperscript{1}LIGM, \'Ecole des Ponts, Univ Gustave Eiffel, CNRS\\
\textsuperscript{2}Inria and DIENS (ENS-PSL, CNRS, Inria) \quad\quad
\textsuperscript{3}Center for Data Science, New York University\\
\href{https://imagine.enpc.fr/~monniert/DTI-Sprites}
{\tt imagine.enpc.fr/\httilde monniert/DTI-Sprites}\\
}

\maketitle

\begin{abstract}
  We present an unsupervised learning framework for decomposing images into layers of 
  automatically discovered object models. Contrary to recent approaches that model image 
  layers with autoencoder networks, we represent them as explicit transformations of a small 
  set of prototypical images. Our model has three main components: (i) a set of object 
  prototypes in the form of learnable images with a transparency channel, which we refer to 
  as sprites; (ii) differentiable parametric functions predicting occlusions and 
  transformation parameters necessary to instantiate the sprites in a given image; (iii) a 
  layered image formation model with occlusion for compositing these instances into complete 
  images including background. By jointly learning the sprites and occlusion/transformation 
  predictors to reconstruct images, our approach not only yields
  accurate layered image decompositions, but also identifies object categories and instance 
  parameters. We first validate our approach by providing results on par with the state of 
  the art on standard multi-object synthetic benchmarks (Tetrominoes, Multi-dSprites, 
  CLEVR6). We then demonstrate the applicability of our model to real images in tasks that 
  include clustering (SVHN, GTSRB), cosegmentation (Weizmann Horse) and object discovery from 
  unfiltered social network images. To the best of our knowledge, our approach is the first 
  layered image decomposition algorithm that learns an explicit and shared concept of object 
  type, and is robust enough to be applied to real images.
\end{abstract}

\begin{figure}[t]
  \begin{subfigure}{\linewidth}
      \centering
      \includegraphics[width=\linewidth]{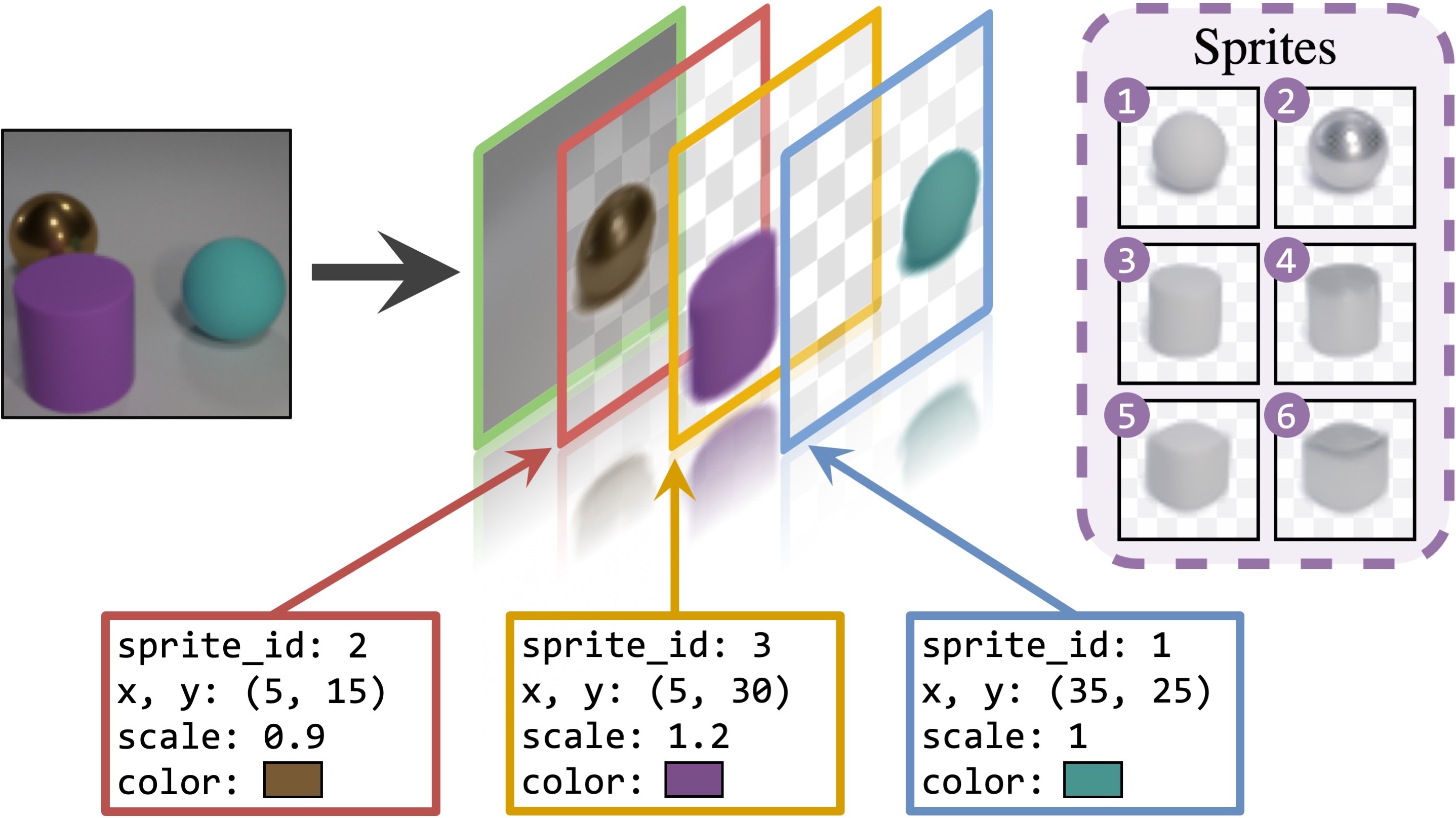}
  \end{subfigure}
  \begin{subfigure}{\linewidth}
      \vspace{.6em}
      \centering
      \includegraphics[width=\linewidth]{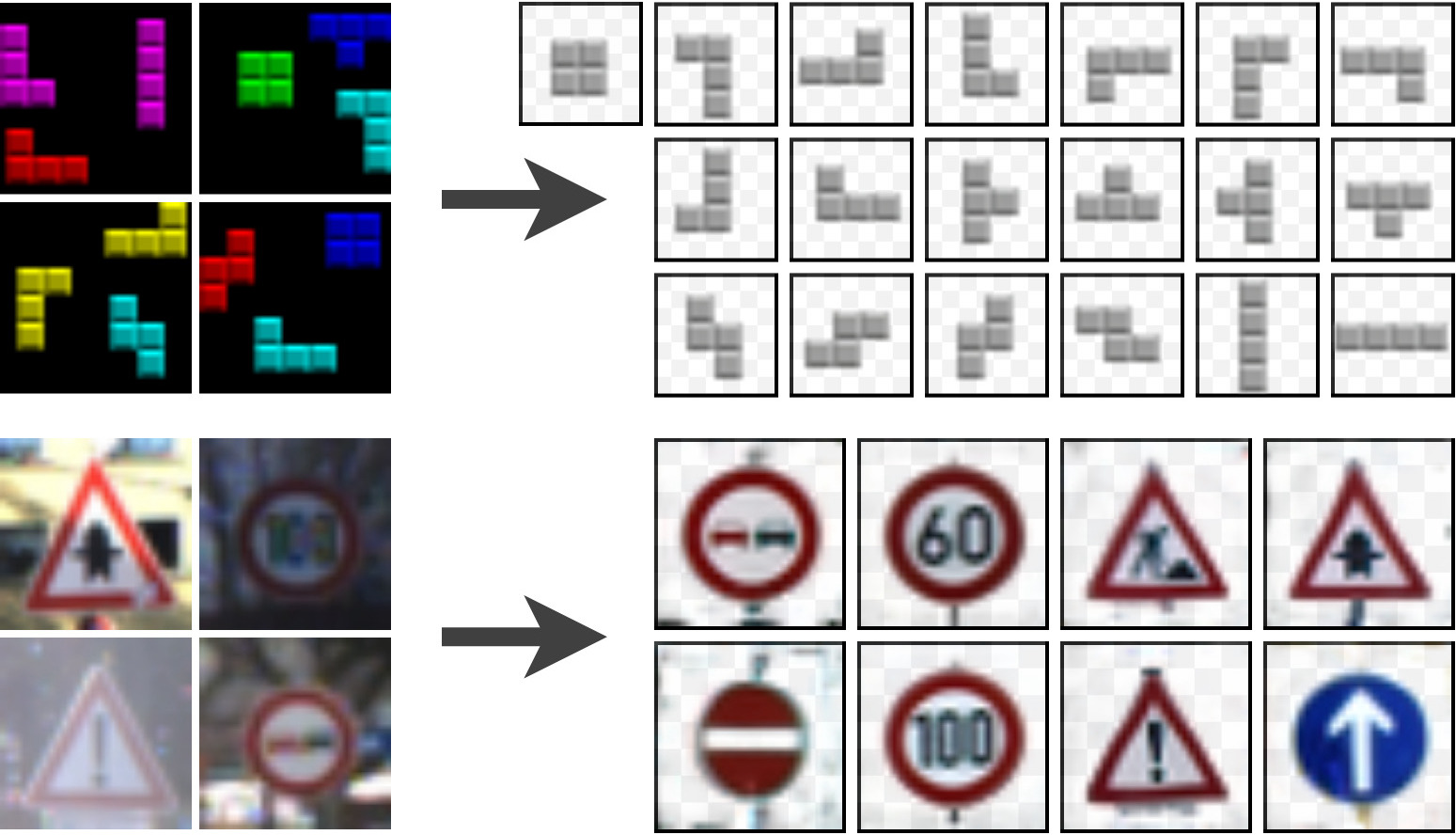}
  \end{subfigure}
  \caption{Our approach learns without supervision to decompose images into layers modeled as 
    transformed instances of prototypical objects called \textit{sprites}. We show an example 
    of decomposition on CLEVR~\cite{johnsonCLEVRDiagnosticDataset2017} (top) and examples of 
    discovered sprites for Tetrominoes~\cite{greffMultiObjectRepresentationLearning2019} and 
    GTSRB~\cite{stallkampManVsComputer2012} (bottom). Transparency is visualized using light 
    gray checkerboards.}
  \label{fig:teaser}
\end{figure}

\section{Introduction}
The aim of this paper is to learn without any supervision a layered decomposition of images, 
where each layer is a transformed instance of a prototypical object. Such an interpretable 
and layered model of images could be beneficial for a plethora of applications like object 
discovery~\cite{eslamiAttendInferRepeat2016, burgessMONetUnsupervisedScene2019}, image 
edition~\cite{wuNeuralSceneDerendering2017, greffMultiObjectRepresentationLearning2019}, 
future frame prediction~\cite{wuLearningSeePhysics2017}, object pose 
estimation~\cite{romaszkoVisionasInverseGraphicsObtainingRich2017} or environment 
abstraction~\cite{anandUnsupervisedStateRepresentation2019, 
linSPACEUnsupervisedObjectOriented2020}. Recent works in this 
direction~\cite{burgessMONetUnsupervisedScene2019, 
greffMultiObjectRepresentationLearning2019, locatelloObjectCentricLearningSlot2020} typically 
learn layered image decompositions by generating layers with autoencoder networks. In 
contrast, we explicitly model them as transformations of a set of prototypical images with 
transparency, which we refer to as {\it sprites}. These sprites are mapped onto their 
instances through geometric and colorimetric transformations resulting in what we call {\it 
object layers}. An image is then assembled from ordered object layers so that each layer 
occludes previous ones in regions where they overlap.  

Our composition model is reminiscent of the classic computer graphics sprite model, popular 
in console and arcade games from the 1980s. While classical sprites were simply placed at 
different positions and composited with a background, we revisit the notion in a spirit 
similar to Jojic and Frey's work on video modeling~\cite{jojicLearningFlexibleSprites2001} by 
using the term in a more generic sense: our sprites can undergo rich geometric 
transformations and color changes.

We jointly learn in an unsupervised manner both the sprites and parametric functions 
predicting their transformations to explain images. This is related to the recent deep 
transformation-invariant (DTI) method designed for clustering by 
Monnier~\etal~\cite{monnierDeepTransformationInvariantClustering2020}.
Unlike this work, however, we handle images that involve a variable number of objects with
limited spatial supports, explained by different transformations and potentially occluding 
each other. This makes the problem very challenging because objects cannot be treated 
independently and the possible number of image compositions is exponential in the number of 
layers.

We experimentally demonstrate in Section~\ref{sec:multi} that our method is on par with the 
state of the art on the synthetic datasets commonly used for image decomposition 
evaluation~\cite{greffMultiObjectRepresentationLearning2019}.  Because our approach 
explicitly models image compositions and object transformations, it also enables us to 
perform simple and controlled image manipulations on these datasets.  More importantly, we 
demonstrate that our model can be applied to real images (Section~\ref{sec:real_img}), where 
it successfully identifies objects and their spatial extent.  For example, we report an 
absolute 5\% increase upon the state of the art on the popular SVHN 
benchmark~\cite{netzer2011reading} and good cosegmentation results on the Weizmann Horse 
database~\cite{borensteinLearningSegment2004}.  We also qualitatively show that our model 
successfully discriminates foreground from background on challenging sets of social network 
images.

\vspace{-1em}
\paragraph{Contributions.} To summarize, we present:
\begin{itemize}
  \item an unsupervised learning approach that explains images as layered compositions of 
    transformed sprites with a background model;
  \item strong results on standard synthetic multi-object benchmarks using the usual instance 
    segmentation evaluation, and an additional evaluation on \textit{semantic} segmentation, 
    which to the best of our knowledge has never been reported by competing methods; and
  \item results on real images for clustering and cosegmentation, which we believe has never 
    been demonstrated by earlier unsupervised image decomposition models.
\end{itemize}
\vspace{0.5em}
\noindent Code and data are available on our project 
\href{https://imagine.enpc.fr/~monniert/DTI-Sprites/}{webpage}.

\begin{figure*}[t]
  \center
  \includegraphics[width=0.99\textwidth]{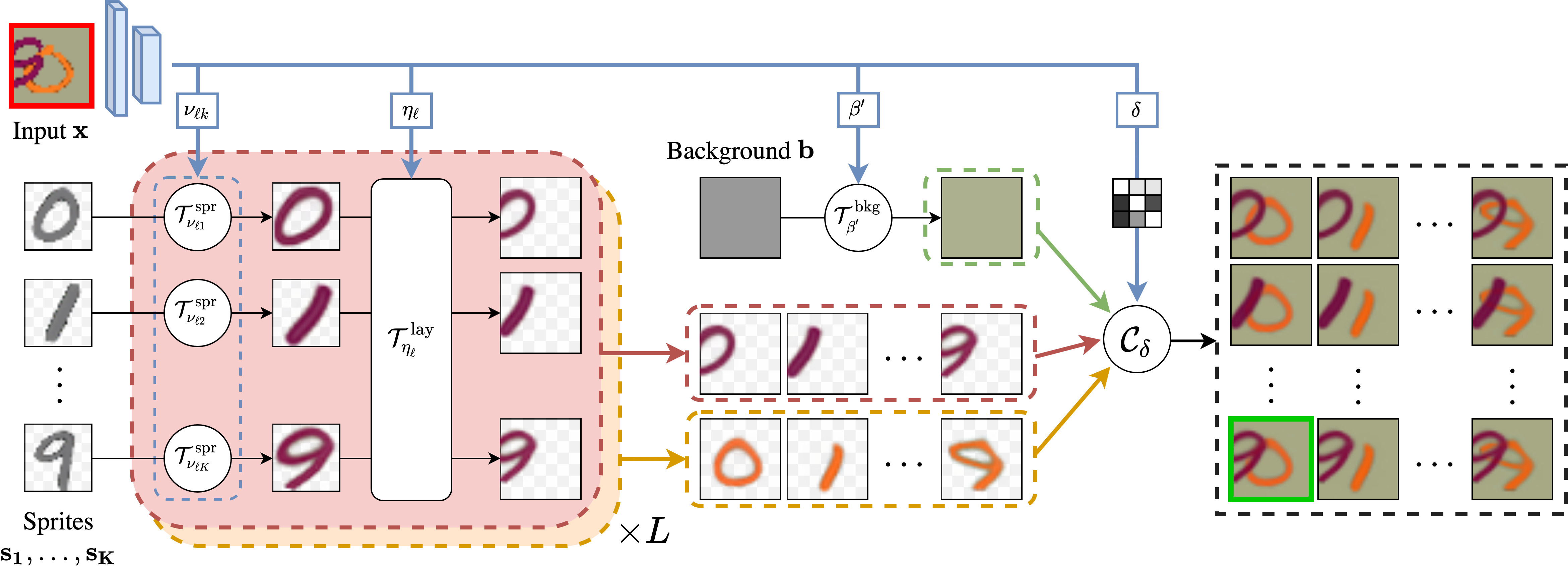}
  \caption{\textbf{Overview.} Given an \red{input image} (highlighted in red) we predict for
    each layer the transformations to apply to the sprites that best reconstruct the input.  
  Transformed sprites and background can be composed into many possible reconstructions given 
a predicted occlusion matrix $\delta$. We introduce a greedy algorithm to select the 
\mygreen{best reconstruction} (highlighted in green).}
  \label{fig:generation}
  \vspace{-0.5em}
\end{figure*}

\section{Related work}\label{sec:relate}

\paragraph{Layered image modeling.} The idea of building images by compositing successive 
layers can already be found in the early work of 
Matheron~\cite{matherongeorgesModeleSequentielPartition1968} introducing the dead leaves 
models, where an image is assembled as a set of templates partially occluding one another and 
laid down in layers. Originally meant for material statistics analysis, this work was 
extended by Lee~\etal~\cite{leeOcclusionModelsNatural2001} to a scale-invariant 
representation of natural images. Jojic and Frey~\cite{jojicLearningFlexibleSprites2001}
proposed to decompose video sequences into layers undergoing spatial modifications - called 
flexible sprites - and demonstrated applications to video editing. Leveraging this idea, Winn 
and Jojic~\cite{winnLOCUSLearningObject2005} introduced LOCUS, a method for learning an 
object model from unlabeled images, and evaluated it for foreground segmentation. Recently, 
approaches~\cite{yangLRGANLayeredRecursive2017, linSTGANSpatialTransformer2018, 
singhFineGANUnsupervisedHierarchical2019, chenUnsupervisedObjectSegmentation2019, 
arandjelovicObjectDiscoveryCopyPasting2019} have used generative adversarial 
networks~\cite{goodfellowGenerativeAdversarialNets2014} to learn layered image compositions, 
yet they are limited to foreground/background modeling. While we also model a layered image 
formation process, we go beyond the simpler settings of image sequences and 
foreground/background separation by decomposing images into multiple objects, each belonging 
to different categories and potentially occluding each other.

\vspace{-1em}
\paragraph{Image decomposition into objects.}
Our work is closely related to a recent trend of works leveraging deep learning to learn 
object-based image decomposition in an unsupervised setting. A first line of works tackles 
the problem from a spatial mixture model perspective where latent variables encoding pixel 
assignment to groups are estimated. Several works from Greff 
\etal~\cite{greffBindingReconstructionClustering2016, greffTaggerDeepUnsupervised2016, 
greffNeuralExpectationMaximization2017} introduce spatial mixtures, model complex pixel 
dependencies with neural networks and use an iterative refinement to estimate the mixture 
parameters.
MONet~\cite{burgessMONetUnsupervisedScene2019} jointly learns a recurrent segmentation 
network and a variational autoencoder (VAE) to predict component mask and appearance.
IODINE~\cite{greffMultiObjectRepresentationLearning2019} instead uses iterative variational 
inference to refine object representations jointly decoded with a spatial broadcast 
decoder~\cite{wattersSpatialBroadcastDecoder2019} as mixture assignments and components.  
Combining ideas from MONet and IODINE, GENESIS~\cite{engelckeGENESISGenerativeScene2020} 
predicts, with an autoregressive prior, object mask representations used to autoencode masked 
regions of the input.  Leveraging an iterative attention 
mechanism~\cite{vaswaniAttentionAllYou2017}, Slot 
Attention~\cite{locatelloObjectCentricLearningSlot2020} produces object-centric 
representations decoded in a fashion similar to IODINE.\ Other related methods 
are~\cite{vansteenkisteRelationalNeuralExpectation2018, vonkugelgenCausalGenerativeScene2020, 
yangLearningManipulateIndividual2020}.  Another set of approaches builds upon the work of 
Eslami \etal introducing AIR~\cite{eslamiAttendInferRepeat2016}, a VAE-based model using 
spatial attention~\cite{jaderbergSpatialTransformerNetworks2015} to iteratively specify 
regions to reconstruct. This notably includes SQAIR~\cite{kosiorekSequentialAttendInfer2018}, 
SPAIR~\cite{crawfordSpatiallyInvariantUnsupervised2019} and the more recent
SPACE~\cite{linSPACEUnsupervisedObjectOriented2020}, which in particular incorporates spatial 
mixtures to model complex backgrounds.  To the best of our knowledge, none of these 
approaches explicitly model object categories, nor demonstrate applicability 
to real images. In contrast, we represent each type of object by a different prototype and 
show results on Internet images.

\vspace{-1em}
\paragraph{Transformation-invariant clustering.} Identifying object categories in an 
unsupervised way can be seen as a clustering problem if each image contains a single object.  
Most recent approaches perform clustering on learned 
features~\cite{hausserAssociativeDeepClustering2018, jiInvariantInformationClustering2019, 
kosiorekStackedCapsuleAutoencoders2019, vangansbekeSCANLearningClassify2020} and do not 
explicitly model the image. In contrast, transformation-invariant clustering explicitly 
models transformations to align images before clustering them. Frey and Jojic first 
introduced this framework~\cite{freyEstimatingMixtureModels1999, 
freyTransformationinvariantClusteringUsing2003} by integrating pixel permutation variables 
within an Expectation-Maximization (EM)~\cite{dempsterMaximumLikelihoodIncomplete1977} 
procedure. Several works~\cite{miller2000learning, learned2005data, cox2008least, 
cox2009least} developed a similar idea for continuous parametric transformations in the 
simpler setting of image alignment, which was later applied again for clustering 
by~\cite{liu2009simultaneous, mattar2012unsupervised, li2016transformation, 
annunziata2019jointly}. Recently, Monnier 
\etal~\cite{monnierDeepTransformationInvariantClustering2020} generalize these ideas to 
global alignments and large-scale datasets by leveraging neural networks to predict spatial 
alignments - implemented as spatial 
transformers~\cite{jaderbergSpatialTransformerNetworks2015} -, color transformations and 
morphological modifications. Also related to ours, 
SCAE~\cite{kosiorekStackedCapsuleAutoencoders2019} leverages the idea of 
capsule~\cite{hintonTransformingAutoEncoders2011} to learn affine-aware image features.  
However, discovered capsules are used as features for clustering and applicability to image 
decomposition into objects has not been demonstrated.

\vspace{-1em}
\paragraph{Cosegmentation and object discovery.} Our method can also be related to 
traditional approaches for object discovery, where the task is to identify and locate objects 
without supervision. A first group of methods~\cite{sivicDiscoveringObjectsTheir2005, 
russellUsingMultipleSegmentations2006, caoSpatiallyCoherentLatent2007, 
sivicUnsupervisedDiscoveryVisual2008} characterizes images as visual words to leverage 
methods from topic modeling and localize objects. Another group of methods aims at computing 
similarities between regions in images and uses clustering models to discover objects. This 
notably includes~\cite{graumanUnsupervisedLearningCategories2006, 
joulinDiscriminativeClusteringImage2010, vicenteObjectCosegmentation2011, 
rubioUnsupervisedCosegmentation2012, joulinMulticlassCosegmentation2012} for cosegmentation 
and~\cite{rubinsteinUnsupervisedJointObject2013, choUnsupervisedObjectDiscovery2015, 
voUnsupervisedImageMatching2019, liGroupWiseDeepObject2019, 
voUnsupervisedMultiobjectDiscovery2020} for object discovery. Although such approaches
demonstrate strong results, they typically use hand-crafted features like saliency measures 
or off-the-shelf object proposal algorithms which are often supervised.  More importantly, 
they do not include an image formation model.

\section{Approach}\label{sec:method}

In this section, we first present our image formation model (Sec.~\ref{sec:gen_model}), then 
describe our unsupervised learning strategy (Sec.~\ref{sec:learning}). 
Figure~\ref{fig:generation} shows an overview of our approach.

\vspace{-1.2em}
\paragraph{Notations.} We write $a_{1:n}$ the ordered set $\{a_1, \ldots, a_n\}$, $\odot$ 
pixel-wise multiplication and use bold notations for images.
Given $N$ colored images $\inp_{1:N}$ of size $H\times W$, we want to learn their 
decomposition into $L$ object layers defined by the instantiations of $K$ sprites. 

\subsection{Image formation model}\label{sec:gen_model}

\paragraph{Layered composition process.} Motivated by early works on layered image
models~\cite{matherongeorgesModeleSequentielPartition1968, jojicLearningFlexibleSprites2001}, 
we propose to decompose an image into $L$ object layers $\obj_{1:L}$ which are overlaid on 
top of each other.  Each object layer $\obj_{\ell}$ is a four-channel image of size $H\times 
W$, three channels correspond to a colored RGB appearance image $\oapp_{\ell}$, and the last 
one $\omask_\ell$ is a transparency or alpha channel over $\oapp_{\ell}$.  Given layers 
$\obj_{1:L}$, we define our image formation process as a recursive composition:
\begin{equation}
  \forall \ell>0,\;\comp_{\ell} = \omask_{\ell} \odot \oapp_{\ell} + (\mathbf{1} - 
  \omask_{\ell}) \odot \comp_{\ell-1},
  \label{eq:recursive}
\end{equation}%
where $\comp_{0} = \mathbf{0}$, and the final result of the composition is $\comp_{L}$. Note 
that this  process explicitly models occlusion: the first layer corresponds to the farthest 
object from the camera, and layer $L$ is the closest, occluding all the others.  In 
particular, we model background by using a first layer with $\omask_1 = 1$. 

Unfolding the recursive process in Eq.~(\ref{eq:recursive}), the layered composition process 
can be rewritten in the compact form:
\begin{equation}
  \cC_\occ(\obj_{1}, \ldots, \obj_{L})= \sum_{\ell=1}^{L} \Big(\prod_{j=1}^{L} (\mathbf{1} - 
  \occ_{j \ell} \omask_j)\Big) \odot \omask_\ell \odot \oapp_\ell,
  \label{eq:compo}
\end{equation}%
where $\occ_{j \ell}=\mathbbm{1}_{[j > \ell]}$ is the indicator function of $j>\ell$. $\occ$ 
is a $L\times L$ binary matrix we call \textit{occlusion matrix}: for given indices $j$ and 
$\ell$, $\occ_{j \ell} = 1$ if layer $j$ occludes layer $\ell$, and $\occ_{j \ell} = 0$ 
otherwise. This gives Eq.~(\ref{eq:compo}) a clear interpretation: each layer appearance 
$\oapp_\ell$ is masked by its own transparency channel $\omask_\ell$ and other layers $j$ 
occluding it, \ie for which $\occ_{j\ell}=1$. Note that we explicitly introduce the 
dependency on $\occ$ in the composition process $\cC_\occ$ because we will later predict it, 
which intuitively corresponds to a layer reordering.

\vspace{-1em}
\paragraph{Sprite modeling.} We model each layer as an explicit transformation of one of $K$ 
learnable sprites $\proto_{1:K}$, which can be seen as prototypes representing the object 
categories. Each sprite $\proto_k$ is a learnable four-channel image of arbitrary size, an 
RGB appearance image $\papp_{k}$ and a transparency channel $\mask_{k}$. To handle variable 
number of objects, we model object absence with an empty sprite $\proto_0 = \mathbf{0}$ added 
to the $K$ sprite candidates and penalize the use of non-empty sprites during 
learning (see Sec.~\ref{sec:learning}). Such modeling assumes we know an upper bound of the 
maximal number of objects, which is rather standard in such a
setting~\cite{burgessMONetUnsupervisedScene2019, greffMultiObjectRepresentationLearning2019, 
locatelloObjectCentricLearningSlot2020}.

Inspired by the recent deep transformation-invariant (DTI) framework designed for 
clustering~\cite{monnierDeepTransformationInvariantClustering2020}, we assume that we have 
access to a family of differentiable transformations $\cT_\beta$ parametrized by $\beta$ - 
\eg an affine transformation with $\beta$ in $\RR^6$ implemented with a spatial 
transformer~\cite{jaderbergSpatialTransformerNetworks2015} - and we model each layer as the 
result of the transformation $\cT_\beta$ applied to one of the $K$ sprites.  We define two 
sets of transformations for a given layer $\ell$: (i) $\tfl_{\parl_\ell}$ the transformations 
parametrized by $\parl_\ell$ and
shared for all sprites in that layer, and (ii)
$\tfp_{\parp_{\ell k}}$ the transformations specific to each sprite and parametrized by 
$\parp_{\ell k}$. More formally, for given layer $\ell$ and sprite $k$ we write:
\begin{equation}
 \cT_{\beta_{\ell k}} (\proto_k) = \tfl_{\parl_\ell} \circ \tfp_{\parp_{\ell k}}(\proto_k),
\end{equation}%
where $\beta_{\ell k} = (\parl_{\ell}, \parp_{\ell k})$ and $\cT_{(\parl_{\ell}, \parp_{\ell 
k})} = \tfl_{\parl_\ell} \circ \tfp_{\parp_{\ell k}}$. 

Although it could be included in $\tfp_{\parp_{\ell k}}$, we separate $\tfl_{\parl_\ell}$ to 
constrain transformations and avoid bad local minima. For example, we use it to model
a coarse spatial positioning so that all sprites in a layer attend to the same object in the 
image. On the contrary, we use $\tfp_{\parp_{\ell k}}$ to model sprite specific deformations, 
such as local elastic deformations.

When modeling background, we consider a distinct set of $K'$ background prototypes 
$\bkg_{1:K'}$, without transparency, and different families of transformations 
$\tfb_{\beta'}$. For simplicity, we write the equations for the case without background and 
omit sprite-specific transformations in the rest of the paper, writing 
$\cT_{\beta_{\ell}}(\proto_k)$ instead of $\cT_{\beta_{\ell k}}(\proto_k)$.\\

\vspace{-0.5em}
To summarize, our image formation model is defined by the occlusion matrix $\occ$, the 
per-layer sprite selection $(k_1, \ldots, k_L)$, the corresponding transformation parameters 
$(\beta_{1}, \ldots, \beta_{L})$, and outputs an image $\inp$ such that:
\begin{equation}
  \inp = \cC_\occ\Big(\cT_{\beta_{1}}(\proto_{k_1}), \ldots, \cT_{\beta_{L}}(\proto_{k_L}
  )\Big).
\end{equation}%
We illustrate our image formation model in Figure~\ref{fig:teaser} and provide a detailed 
example in Figure~\ref{fig:generation}.

\subsection{Learning}\label{sec:learning}

We learn our image model without any supervision by minimizing the objective function:%
\begin{multline}
  \cL(\proto_{1:K}, \dpg_{1:L}, \dpp) = \sum_{i=1}^N\; \min_{k_1,\ldots,k_L} \Big( \lambda 
    \sum_{j=1}^L \mathbbm{1}_{[k_j \neq 0]} + \\ \Big\|\inp_i - 
    \cC_{\dpp(\inp_i)}\scaleto{\Big(\cT_{{\dpg_{1}}(\inp_i)}(\proto_{k_{1}}),\ldots,
  \cT_{{\dpg_{L}}(\inp_i)}(\proto_{k_L})\Big)}{17pt}\Big\|^2_2 \Big),
  \label{eq:loss}
\end{multline}%
where $\proto_{1:K}$ are the sprites, $\phi_{1:L}$ and $\psi$ are neural networks
predicting the transformation parameters and occlusion matrix for a given image $\inp_i$, 
$\lambda$ is a scalar hyper-parameter and $\mathbbm{1}_{[k_j \neq 0]}$ is the indicator 
function of $k_j \neq 0$. The first sum is over all images in the database, the minimum 
corresponds to the selection of the sprite used for each layer and the second sum counts the 
number of non-empty sprites. If $\lambda>0$, this loss encourages reconstructions using the 
minimal number of non-empty sprites. In practice, we use $\lambda = 10^{-4}$.

Note the similarity between our loss and the gradient-based 
adaptation~\cite{bottouConvergencePropertiesKmeans1995} of the K-means 
algorithm~\cite{macqueenMethodsClassificationAnalysis1967} where the squared Euclidean 
distance to the closest prototype is minimized, as well as with its transformation-invariant 
version~\cite{monnierDeepTransformationInvariantClustering2020} including
neural networks modeling transformations. In addition to the layered composition model 
described in the previous section, the main two differences with our model are the joint 
optimization over $L$ sprite selections and the occlusion modeling that we discuss next.

\vspace{-1em}
\paragraph{Sprites selection.} Because the minimum in Eq.~(\ref{eq:loss}) is taken over the 
$(K+1)^L$ possible selections leading to as many reconstructions, an exhaustive search over 
all combinations quickly becomes impossible when dealing with many objects and layers.  Thus, 
we propose an iterative greedy algorithm to estimate the minimum, described in 
Algorithm~\ref{algo:greedy} and used when $L > 2$. While the solution it provides is of 
course not guaranteed to be optimal, we found it performs well in practice.
At each iteration, we proceed layer by layer and iteratively select for each layer the sprite 
$k_\ell$ minimizing the loss, keeping all other object layers fixed. This reduces the number 
of reconstructions to perform to $T \times (K+1) \times L$. In practice, we have observed 
that convergence is reached after 1 iteration for Tetrominoes and 2-3 iterations for 
Multi-dSprites and CLEVR6, so we have respectively used $T = 1$ and $T = 3$ in these 
experiments. We experimentally show in our ablation study presented in Sec.~\ref{sec:multi} 
that this greedy approach yields performances comparable to an exhaustive search when 
modeling small numbers of layers and sprites.

\newlength{\textfloatsepsave}
\setlength{\textfloatsepsave}{\textfloatsep}
\setlength{\textfloatsep}{0pt}
\begin{algorithm}[t]
  \caption{Greedy sprite selection.}
  \label{algo:greedy}
  \setlength{\abovedisplayskip}{0pt}
  \setlength{\belowdisplayskip}{0pt}
  \KwIn{image $\inp$, occlusion $\delta$, $(K+1)\times L$ object layers candidates 
  $\cT_{{\dpg_{\ell}}(\inp)}(\proto_{k})$, steps $T$}
  \KwOut{sprite indices $(k_1, \ldots, k_L)$}
  \KwInit{$\forall \ell \in \{1,\ldots,L\},\;  k_\ell \gets {0}, \obj_\ell \gets \mathbf{0}$}
  \For(\hfill{\textit{\# iterations}}){$t = 1,\ldots,T$}{%
    \For(\hfill{\textit{\# loop on layers}}){$\ell = 1,\ldots,L$}{%
      $k_\ell \gets \min_k\, \Big[\lambda \mathbbm{1}_{[k \neq 0]}\;+$ \\ ~~~~~~~~~ 
        $\|\inp-\cC_\occ(\obj_{1:\ell-1},\, \cT_{{\dpg_{\ell}}(\inp)}(\proto_{k}),\,
      \obj_{\ell+1:L})\|^2_2 \Big]$\\
  $\obj_\ell \gets \cT_{{\dpg_{\ell}}(\inp)}(\proto_{k_\ell})$
  }
  }
  \KwRet{$k_1,\ldots, k_L$}
\end{algorithm}

\vspace{-1em}
\paragraph{Occlusion modeling.} Occlusion is modeled explicitly in our composition process 
defined in Eq.~(\ref{eq:compo}) since $\obj_1, \ldots, \obj_L$ are ranked by depth.  However, 
we experimentally observed that layers learn to specialize to different regions in the image.  
This seems to correspond to a local minimum of the loss function, and the model does not 
manage to reorder the layers to predict the correct occlusion. Therefore, we relax the model 
and predict an occlusion matrix $\occ= \dpp(\inp)\in [0, 1]^{L\times L}$ instead of keeping 
it fixed. More precisely, for each image $\inp$ we predict $\frac{1}{2}L(L-1)$ values using a 
neural network followed by a sigmoid function. These values are then reshaped to a lower 
triangular $L \times L$ matrix with zero diagonal, and the upper part is computed by symmetry 
such that: $\forall i<j,\, \occ_{i j} = 1 - \occ_{j i}$. While such predicted occlusion 
matrix is not binary and does not directly translate into a layer reordering, it still allows 
us to compute a composite image using Eq.~(\ref{eq:compo}) and the masks associated to each 
object.  Note that such a matrix could model more complex occlusion relationships such as 
non-transitive ones. At inference, we simply replace $\delta_{ij}$ by $\delta_{ij} > 0.5$ to 
obtain binary occlusion relationships. We also tried computing the closest matrix 
corresponding to a true layer reordering and obtained similar results. Note that when we use 
a background model, its occlusion relationships are fixed, \ie $\forall j>1, \occ_{j 1} = 1$.

\vspace{-1em}
\paragraph{Training details.} Two elements of our training strategy are crucial to the 
success of learning. First, following~\cite{monnierDeepTransformationInvariantClustering2020} 
we adopt a curriculum learning of the transformations, starting by the simplest ones. Second, 
inspired by Tieleman~\cite{tielemanOptimizingNeuralNetworks2014} and 
SCAE~\cite{kosiorekStackedCapsuleAutoencoders2019}, we inject uniform noise in the masks in 
such a way that masks are encouraged to be binary (see supplementary for details). This 
allows us to resolve the ambiguity that would otherwise exist between the color and alpha 
channels and obtain clear masks. We provide additional details about networks' architecture, 
computational cost, transformations used and implementation in the supplementary material.

\section{Experiments}\label{sec:exp}

\setlength{\textfloatsep}{\textfloatsepsave}
Assessing the quality of an object-based image decomposition model is ambiguous and 
difficult, and downstream applications on synthetic multi-object benchmarks such 
as~\cite{multiobjectdatasets19} are typically used as evaluations. Thus, recent approaches 
(\eg~\cite{burgessMONetUnsupervisedScene2019, greffMultiObjectRepresentationLearning2019, 
engelckeGENESISGenerativeScene2020, locatelloObjectCentricLearningSlot2020}) first evaluate 
their ability to infer spatial arrangements of objects through quantitative performances for 
object instance discovery.  The knowledge of the learned concept of object is then evaluated 
qualitatively through convincing object-centric image 
manipulation~\cite{burgessMONetUnsupervisedScene2019, 
greffMultiObjectRepresentationLearning2019}, occluded region 
reconstructions~\cite{burgessMONetUnsupervisedScene2019, 
locatelloObjectCentricLearningSlot2020} or realistic generative 
sampling~\cite{engelckeGENESISGenerativeScene2020}. None of these approaches explicitly model 
categories for objects and, to the best of our knowledge, their applicability is limited to 
synthetic imagery only.

In this section, we first evaluate and analyse our model on the standard
multi-object synthetic benchmarks (Sec.~\ref{sec:multi}). Then, we demonstrate that our 
approach can be applied to real images
(Sec.~\ref{sec:real_img}). We use the 2-layer version of our model to perform clustering 
(\ref{sec:clus}), cosegmentation (\ref{sec:coseg}), as well as qualitative object discovery 
from unfiltered web image collections (\ref{sec:web}). 

\subsection{Multi-object synthetic benchmarks}\label{sec:multi}

\paragraph{Datasets and evaluation.} 
Tetrominoes~\cite{greffMultiObjectRepresentationLearning2019} is a 60k dataset generated by 
placing three Tetrominoes without overlap in a $35\times35$ image.  There is a total of 19 
different Tetrominoes (counting discrete rotations).  
Multi-dSprites~\cite{multiobjectdatasets19} contains 60k images of size $64\times64$ with 2 
to 5 objects sampled from a set of 3 different shapes: ellipse, heart, square.  
CLEVR6~\cite{johnsonCLEVRDiagnosticDataset2017, greffMultiObjectRepresentationLearning2019} 
contains 34,963 synthetically generated images of size $128\times128$. Each image is composed 
of a variable number of objects (from 3 to 6), each sampled from a set of 6 categories - 3 
different shapes (sphere, cylinder, cube) and 2 materials (rubber or metal) - and randomly 
rendered. We thus train our method using one sprite per object category and as many layers as 
the maximum number of objects per image, with a background layer when necessary.  Following
standard practices~\cite{greffMultiObjectRepresentationLearning2019, 
locatelloObjectCentricLearningSlot2020}, we evaluate object instance segmentation on 320 
images by averaging over all images the Adjusted Ranked Index (ARI) computed using 
ground-truth foreground pixels only (ARI-FG in our tables). Note that because background 
pixels are filtered, ARI-FG strongly favors methods 
like~\cite{greffMultiObjectRepresentationLearning2019, 
locatelloObjectCentricLearningSlot2020} which oversegment objects or do not discriminate 
foreground from background. To limit the penalization of our model which explicitly models 
background, we reassign predicted background pixels to the closest object layers before 
computing this metric. However, we argue that foreground/background separation is crucial for 
any downstream applications and also advocate the use of a true ARI metric computed on all 
pixels (including background) which we include in our results. In addition, we think that the 
knowledge of object category should be evaluated and include quantitative results for 
unsupervised semantic segmentation in the supplementary material.

\vspace{-1em}
\paragraph{Results.} Our results are compared quantitatively to state-of-the-art approaches 
in Table~\ref{tab:multiobj}. On Multi-dSprites, an outlier run out of 5 was automatically 
filtered based on its high reconstruction loss compared to the others. Our method obtains 
results on par with the best competing methods across all benchmarks.  While our approach is 
more successful on benchmarks depicting 2D scenes, it still provides good results on CLEVR6 
where images include 3D effects. We provide our results using the real ARI metric which we 
believe to be more interesting as it is not biased towards oversegmenting methods. While this 
measure is not reported by competing methods, a CLEVR6 decomposition example shown in 
official IODINE 
implementation\footnote{\href{https://github.com/deepmind/deepmind-research/blob/master/iodine}
{https://github.com/deepmind/deepmind-research/blob/master/iodine}}
gives a perfect 100\% ARI-FG score but reaches 20\% in terms of ARI.

\begin{table}[t]
  \renewcommand{\arraystretch}{1}
  \addtolength{\tabcolsep}{-4pt}
  \caption{\textbf{Multi-object instance discovery.} Following standard practices, we report 
  ARI-FG (ARI on foreground pixels only) averaged over 5 runs. We also report our results 
with the real ARI, a metric we advocate for future comparisons. We mark results 
($\vartriangle$) where one outlier run is filtered out. }
  \vspace{-.5em}
  \centering
  \small
  \begin{tabular}{@{}lccccc@{}} \toprule
  Method & Metric & Tetrominoes & Multi-dSprites & CLEVR6\\
  \midrule
  MONet~\cite{burgessMONetUnsupervisedScene2019} &ARI-FG& - & 90.4 $\pm$ 0.8 &  96.2 $\pm$ 
  0.6\\
  IODINE~\cite{greffMultiObjectRepresentationLearning2019} &ARI-FG& 99.2 $\pm$ 0.4 &  76.7 
  $\pm$ 5.6 & \bf 98.8 $\pm$ 0.0 \\
  Slot Att.~\cite{locatelloObjectCentricLearningSlot2020} & ARI-FG &
  \underline{99.5$^{\vartriangle}\pm$ 0.2} & \underline{91.3 $\pm$ 0.3} & \bf 98.8 $\pm$ 
  0.3\\

  \textbf{Ours} & ARI-FG & \bf 99.6 $\pm$ 0.2 &\bf 92.5$^{\vartriangle} \pm$ 0.3 & 
  \underline{$97.2 \pm 0.2$}\\
  \midrule
  \textbf{Ours} & ARI & 99.8 $\pm$ 0.1 & 95.1$^{\vartriangle} \pm$  0.1& 90.7 $\pm$ 0.1\\
  \bottomrule
  \end{tabular}
  \label{tab:multiobj}
  \vspace{-0em}
\end{table}

\begin{figure}[t]
  \center
  \includegraphics[width=\linewidth]{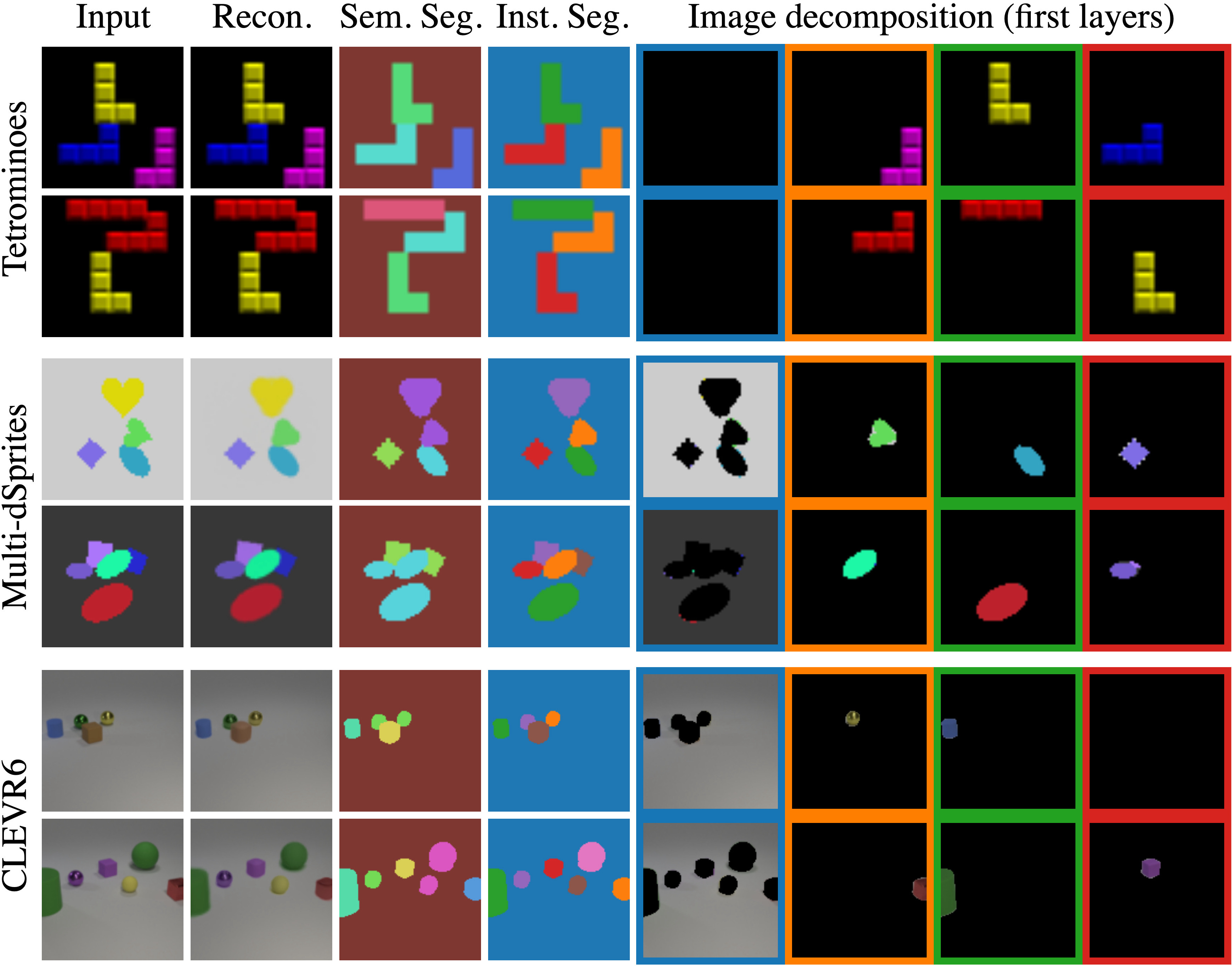}
  \caption{\textbf{Multi-object discovery.} From left to right, we show inputs, 
  reconstructions, semantic (each color corresponds to a different sprite) and instance 
segmentations, and first decomposition layers colored w.r.t.\ their instance mask.}
\label{fig:instance_seg}
\vspace{-0.3em}
\end{figure}

Compared to all competing methods, our approach explicitly models categories for objects. In 
particular, it is able to learn prototypical images that can be associated to each object 
category. The sprites discovered from CLEVR6 and Tetrominoes are shown in 
Fig.~\ref{fig:teaser}. Note how learned sprites on Tetrominoes are sharp and how we can 
identify material in CLEVR6 by learning two different sprites for each shape.

\begin{figure}[t]
  \center
  \includegraphics[width=\linewidth]{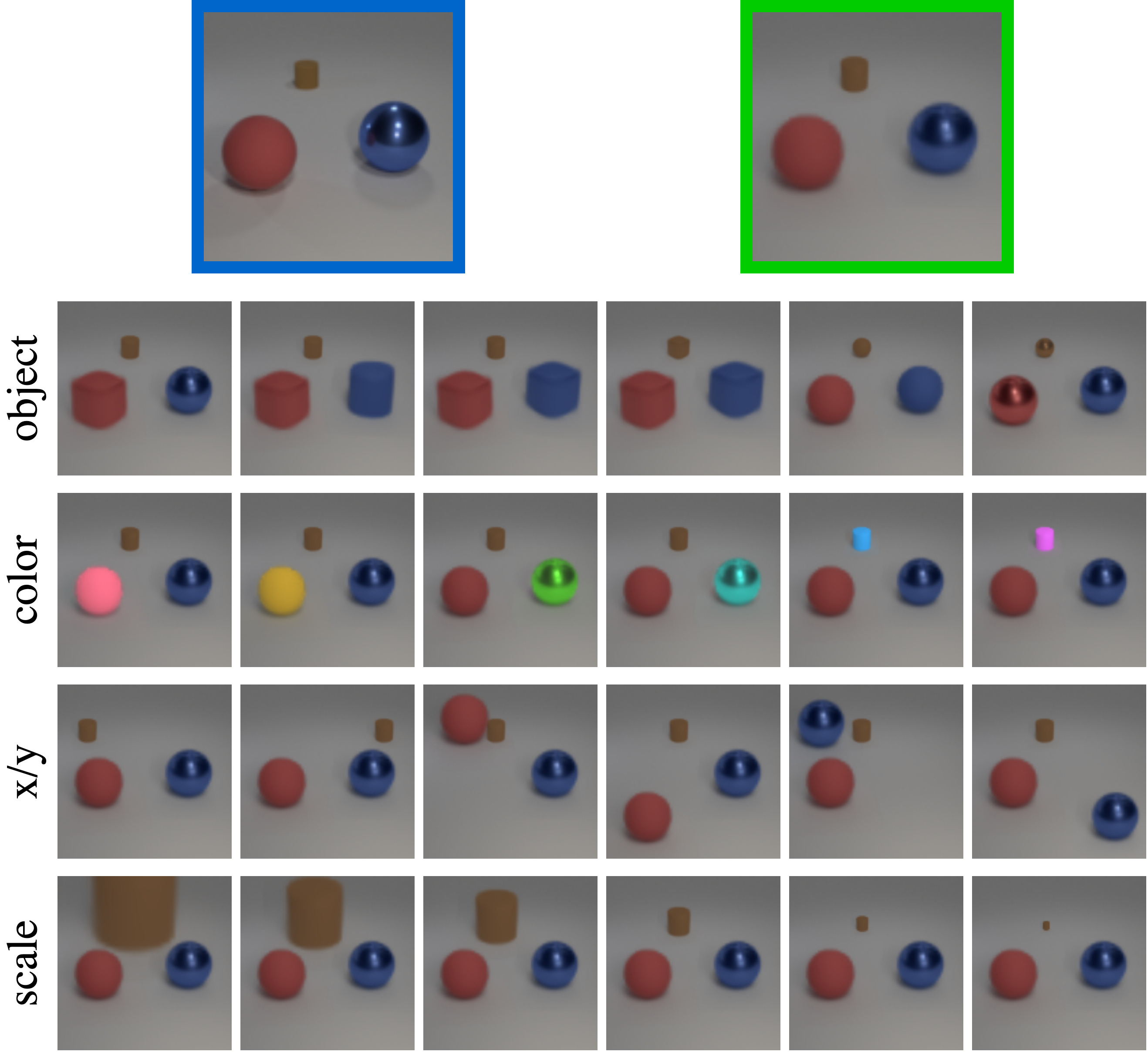}
  \caption{\textbf{Object-centric image manipulation}. Given a \myblue{query image} (top 
    left) from CLEVR6~\cite{johnsonCLEVRDiagnosticDataset2017}, we show the \mygreen{closest 
  reconstruction} (top right) and several image manipulations (next four rows). From top to 
bottom, we respectively use different sprites, change the objects color, vary their positions 
and modify the scale.}
  \label{fig:rendering}
  \vspace{-0.3em}
\end{figure}

In Fig.~\ref{fig:instance_seg}, we show some qualitative results obtained on the three 
benchmarks. Given sample images, we show from left to right the final reconstruction, 
semantic segmentation (evaluated quantitatively in the supplementary material) where each 
color corresponds to a different sprite, instance segmentation, and the first four layers of 
the image decomposition. Note how we manage to successfully predict occlusions, model 
variable number of objects, separate the different instances, as well as identify the object 
categories and their spatial extents.  More random decomposition results are shown in the 
supplementary and
on our \href{https://imagine.enpc.fr/~monniert/DTI-Sprites}{webpage}.

Compared to other approaches which typically need a form of supervision to interpret learned 
representations as object visual variations, our method has the advantage to give a direct 
access to the object instance parameters, enabling us to directly manipulate them in images.  
In Fig.~\ref{fig:rendering}, we show different object-centric image manipulations such as
objects swapping as well as color, position and scale variations. Note that we are also able 
to render out of distribution instances, like the pink sphere or the gigantic cylinder.

\begin{table}[t]
  \renewcommand{\arraystretch}{1}
  \addtolength{\tabcolsep}{-4pt}
  \caption{\textbf{Ablation study.} Results are averaged over 5 runs.}
  \vspace{-.5em}
  \centering
  \small
  \begin{tabular}{@{}llcc@{}} \toprule
  Dataset & Model & ARI-FG & ARI\\
  \midrule
  Multi-dSprites2 & Full &\bf 95.5 $\pm$ 2.1 & 95.2 $\pm$ 1.9\\
  & w/o greedy algo. & 94.4 $\pm$ 2.7 &\bf 95.9 $\pm$ 0.3\\
  \midrule
  Multi-dSprites & Full &\bf 91.5 $\pm$ 2.2 &\bf 95.0 $\pm$ 0.3\\
  & w/o occ.\ prediction & 85.7 $\pm$ 2.2 & 94.2 $\pm$ 0.2\\
  \midrule
  Tetrominoes & Full &\bf 99.6 $\pm$ 0.2 & \bf 99.8 $\pm$ 0.1\\
  & w/o shared transfo. & 95.3 $\pm$ 3.7 & 82.6 $\pm$ 12.2\\
  \bottomrule
  \end{tabular}
  \label{tab:ablation}
  \vspace{-.5em}
\end{table}

\vspace{-1em}
\paragraph{Ablation study.} We analyze the main components of our model in 
Table~\ref{tab:ablation}. For computational reasons, we evaluate our greedy algorithm on 
Multi-dSprites2 - the subset of Multi-dSprites containing only 2 objects - and show 
comparable performances to an exhaustive search over all combinations. Occlusion prediction 
is evaluated on Multi-dSprites which contains many occlusions. Because our model with fixed 
occlusion does not manage to reorder the layers, performances are significantly better when 
occlusion is learned. Finally, we compare results obtained on Tetrominoes when modeling 
sprite-specific transformations only, without shared ones, and show a clear gap between the 
two settings. We provide analyses on the effects of $K$ and $\lambda$ in the supplementary.

\vspace{-1em}
\paragraph{Limitations.} Our optimization model can be stuck in local minima. A typical 
failure mode on Multi-dSprites can be seen in the reconstructions in 
Fig.~\ref{fig:instance_seg} where a triangular shape is learned instead of the heart. This 
sprite can be aligned to a target heart shape using three different equivalent rotations, and 
our model does not manage to converge to a consistent one. This problem could be overcome by 
either modeling more sprites, manually computing reconstructions with different discrete 
rotations, or guiding transformation predictions with supervised sprite transformations.

\subsection{Real image benchmarks}\label{sec:real_img}

\subsubsection{Clustering}\label{sec:clus}

\paragraph{Datasets.}  We evaluate our model on two real image clustering datasets using 2 
layers, one for the background and one for the foreground object.  
SVHN~\cite{netzer2011reading} is a standard clustering dataset composed of digits extracted 
from house numbers cropped from Google Street View images. Following standard 
practices~\cite{huLearningDiscreteRepresentations2017, 
kosiorekStackedCapsuleAutoencoders2019, monnierDeepTransformationInvariantClustering2020}, we 
evaluate on the labeled subset (99,289 images), but also use 530k unlabeled extra samples for 
training.  We also report results on traffic sign images using a balanced subset of the GTSRB 
dataset~\cite{stallkampManVsComputer2012} which we call \mbox{GTSRB-8}{}. We selected classes 
with 1000 to 1500 instances in the training split, yielding 8 classes and 10,650 images which 
we resize to $28\times 28$.

\vspace{-1em}
\paragraph{Results.} We compare our model to state-of-the-art methods in Table~\ref{tab:clus} 
using global clustering accuracy, where the
cluster-to-class mapping is computed using the Hungarian 
algorithm~\cite{kuhnHungarianMethodAssignment1955}. We train our 2-layer model with as many 
sprites as classes and a single background prototype. On both benchmarks, our approach 
provides competitive results.  In particular, we improve state of the art on the standard 
SVHN benchmark by an absolute 5\% increase. 

\begin{figure}[t]
  \centering
  \includegraphics[width=\linewidth]{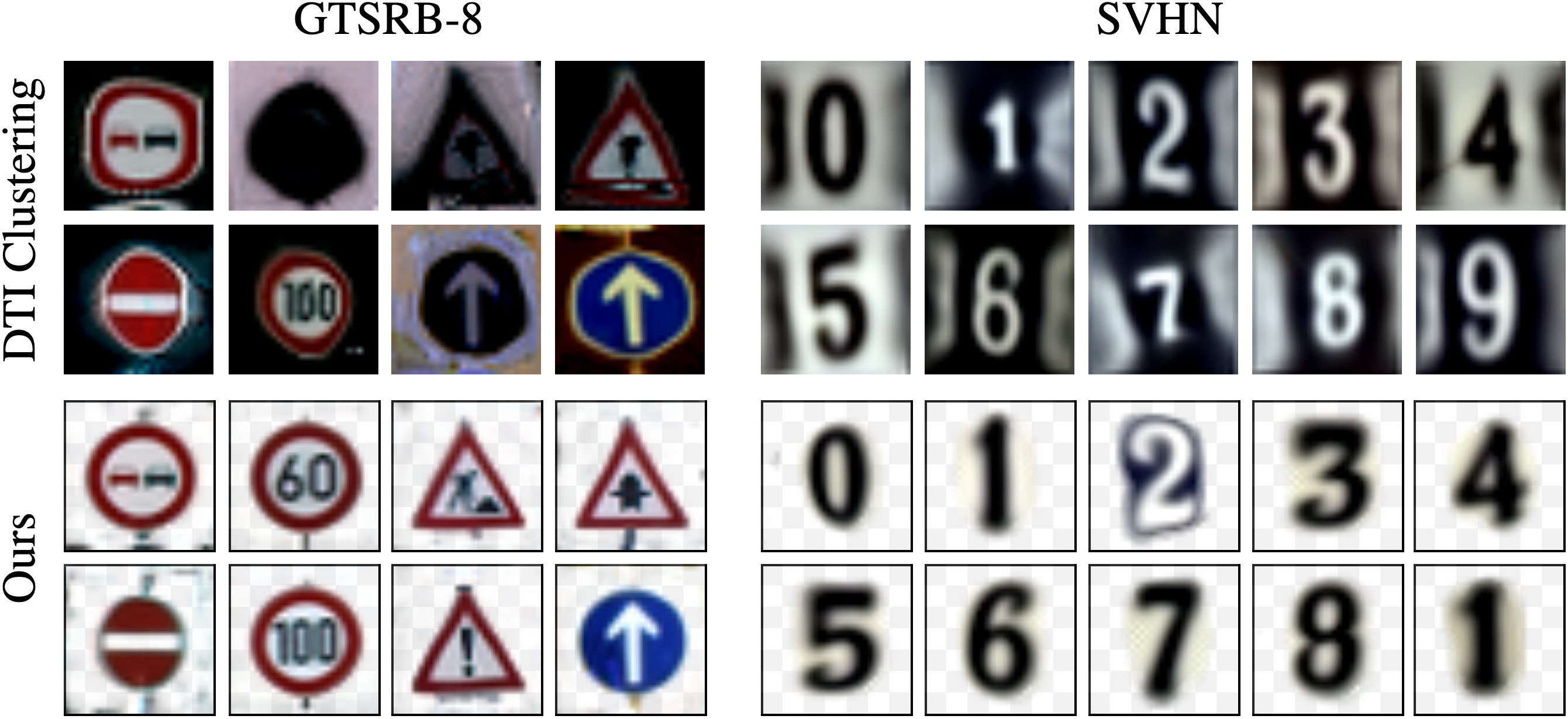}
  \caption{\textbf{Qualitative clustering results.} We compare prototypes learned using
    DTI-Clustering and our discovered sprites on GTSRB-8 (left) and SVHN (right).}
  \label{fig:recons}
  \vspace{-.7em}
\end{figure}

Similar to DTI-Clustering, our method performs clustering in pixel-space exclusively and has 
the advantage of providing interpretable results. Figure~\ref{fig:recons} shows learned 
sprites on the GTSRB-8 and SVHN datasets and compares them to prototypes learned with 
DTI-Clustering. Note in particular the sharpness of the discovered GTSRB-8 sprites.

\begin{table}[t]
  \renewcommand{\arraystretch}{1}
  \addtolength{\tabcolsep}{-4pt}
  \caption{\textbf{Clustering comparisons.} We report average clustering accuracy.  We mark 
  methods we ran ourselves with official implementations ($\star$), use data augmentation 
($\triangledown$) or ad-hoc representations ($\dagger$ for GIST, $\ddagger$ for Sobel 
filters).}
  \vspace{-.5em}
  \centering
  \small
  \begin{tabular}{@{}lccc@{}} \toprule
  Method & Runs & GTSRB-8 & SVHN\\
  \midrule
  \multicolumn{4}{l}{\textit{Clustering on learned features}} \\
  \quad ADC~\cite{hausserAssociativeDeepClustering2018} & 20 & - & 38.6$^{\triangledown}$\\
  \quad SCAE~\cite{kosiorekStackedCapsuleAutoencoders2019} & 5 & - & 55.3$^{\ddagger}$\\
  \quad IMSAT~\cite{huLearningDiscreteRepresentations2017} & 12 & 
  26.9$^{\triangledown\star}$& 57.3$^{\triangledown\dagger}$\\
  \quad SCAN~\cite{vangansbekeSCANLearningClassify2020} & 5 & \bf 
  90.4$^{\triangledown\star}$& 54.2$^{\triangledown\star}$\\

  \midrule
  \multicolumn{4}{l}{\textit{Clustering on pixel values}}\\
  \quad DTI-Clustering~\cite{monnierDeepTransformationInvariantClustering2020}
  & 10 & 54.3$^{\star}$ &\underline{57.4}\\
  \quad \textbf{Ours} & 10 & \underline{89.4} &\bf 63.1\\
  \bottomrule
  \end{tabular}
  \label{tab:clus}
  \vspace{-.5em}
\end{table}

\subsubsection{Cosegmentation}\label{sec:coseg}

\paragraph{Dataset.} We use the Weizmann Horse database~\cite{borensteinLearningSegment2004} 
to evaluate quantitatively the quality of our masks. It is composed of 327 side-view horse 
images resized to $128 \times 128$. Although relatively simple compared to more recent 
cosegmentation datasets, it presents significant challenges compared to previous
synthetic benchmarks because of the diversity of both horses and backgrounds. The dataset was 
mainly used by classical (non-deep) methods which were trained and evaluated on 30 images for 
computational reasons while we train and evaluate on the full set.

\vspace{-1em}
\paragraph{Results.} We compare our 2-layer approach with a single sprite to classical
cosegmentation methods in Table~\ref{tab:coseg} and report segmentation accuracy - mean \% of 
pixels correctly classified as foreground or background - averaged over 5 runs. Our results 
compare favorably to these classical approaches. Although more recent approaches could 
outperform our method on this dataset, we argue that obtaining performances on par with such 
competing methods is already a strong result for our layered image decomposition model.

\begin{figure}[t]
  \center
  \includegraphics[width=\linewidth]{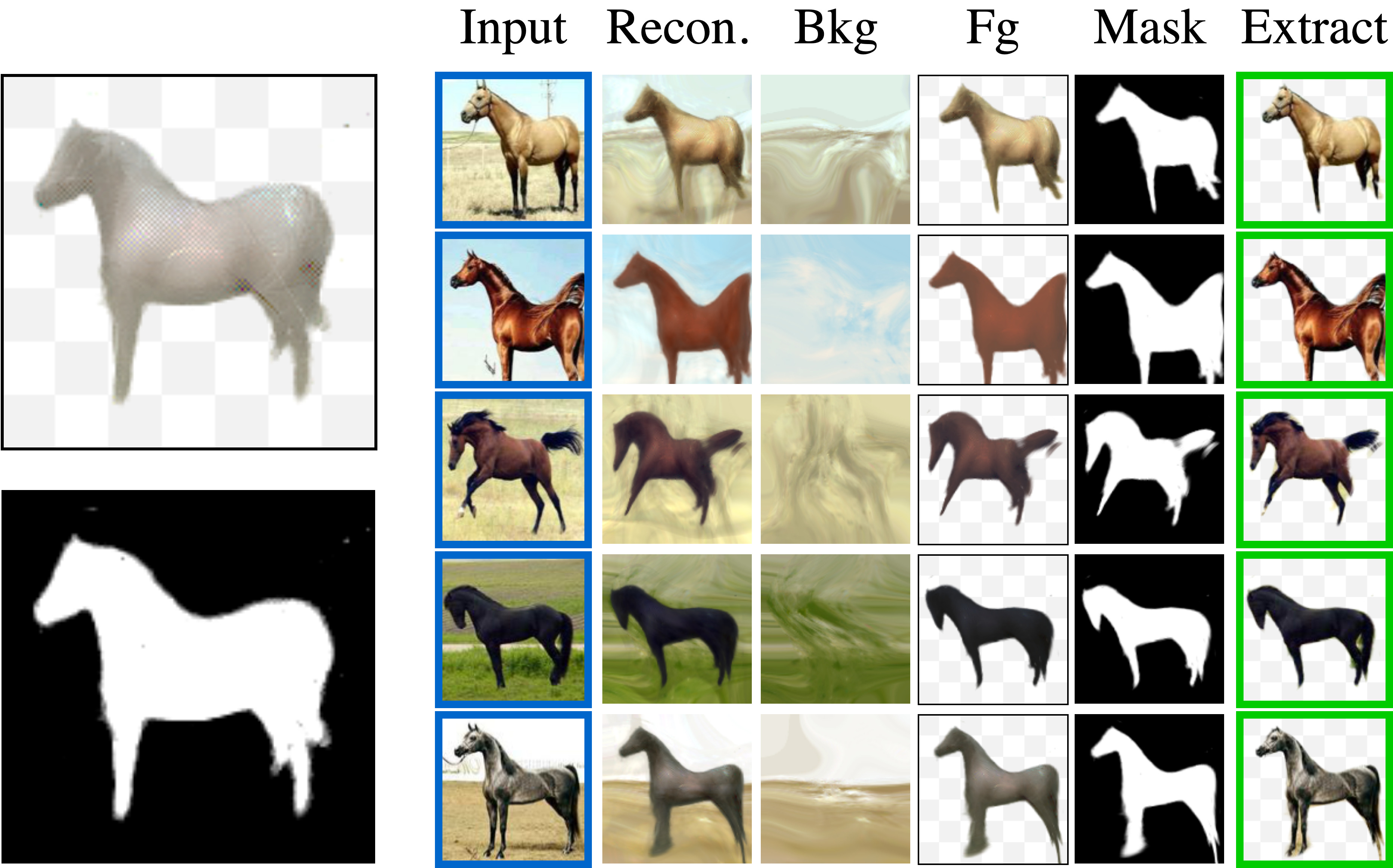}
  \caption{\textbf{Qualitative cosegmentation results.} Sprite and mask (left) learned from 
  Weizmann Horse~\cite{borensteinLearningSegment2004} and some result examples (right) giving 
  for each \myblue{input}, its reconstruction, the layered composition and  
\mygreen{extracted foreground}.}
  \label{fig:horse}
  \vspace{-.7em}
\end{figure}

We present in Fig.~\ref{fig:horse} some visual results of our approach. First, the discovered 
sprite clearly depicts a horse shape and its masks is sharp and accurate.  Learning such an 
interpretable sprite from this real images collection is already interesting and validates 
that our sprite-based modeling generalizes to real images.  Second, although the 
transformations modeled are quite simple (a combination of color and spatial 
transformations), we demonstrate good reconstructions and decompositions, yielding accurate 
foreground extractions.

\begin{table}[t]
  \renewcommand{\arraystretch}{1}
  \addtolength{\tabcolsep}{-2pt}
  \caption{\textbf{Weizmann Horse cosegmentation comparisons.}}
  \vspace{-.7em}
  \centering
  \small
  \begin{tabular}{@{}lcccccc@{}} \toprule
   Method & \cite{rubioUnsupervisedCosegmentation2012} & 
   \cite{joulinDiscriminativeClusteringImage2010} & 
   \cite{lattariUnsupervisedCosegmentationBased2015} &  
   \cite{changFromCosaliencyToCosegmentation2011} &\cite{yuUnsupervisedCosegmentation2014} & 
   Ours\\
  \midrule
   Accuracy (\%) & 74.9 & 80.1 & 84.6 &  86.4 & \underline{87.6} &\bf 87.9\\
  \bottomrule
  \end{tabular}
  \label{tab:coseg}
  \vspace{-.5em}
\end{table}

\subsubsection{Unfiltered web image collections}\label{sec:web}

\begin{figure}[t]
  \begin{subfigure}{\linewidth}
    \centering
    \includegraphics[width=\linewidth]{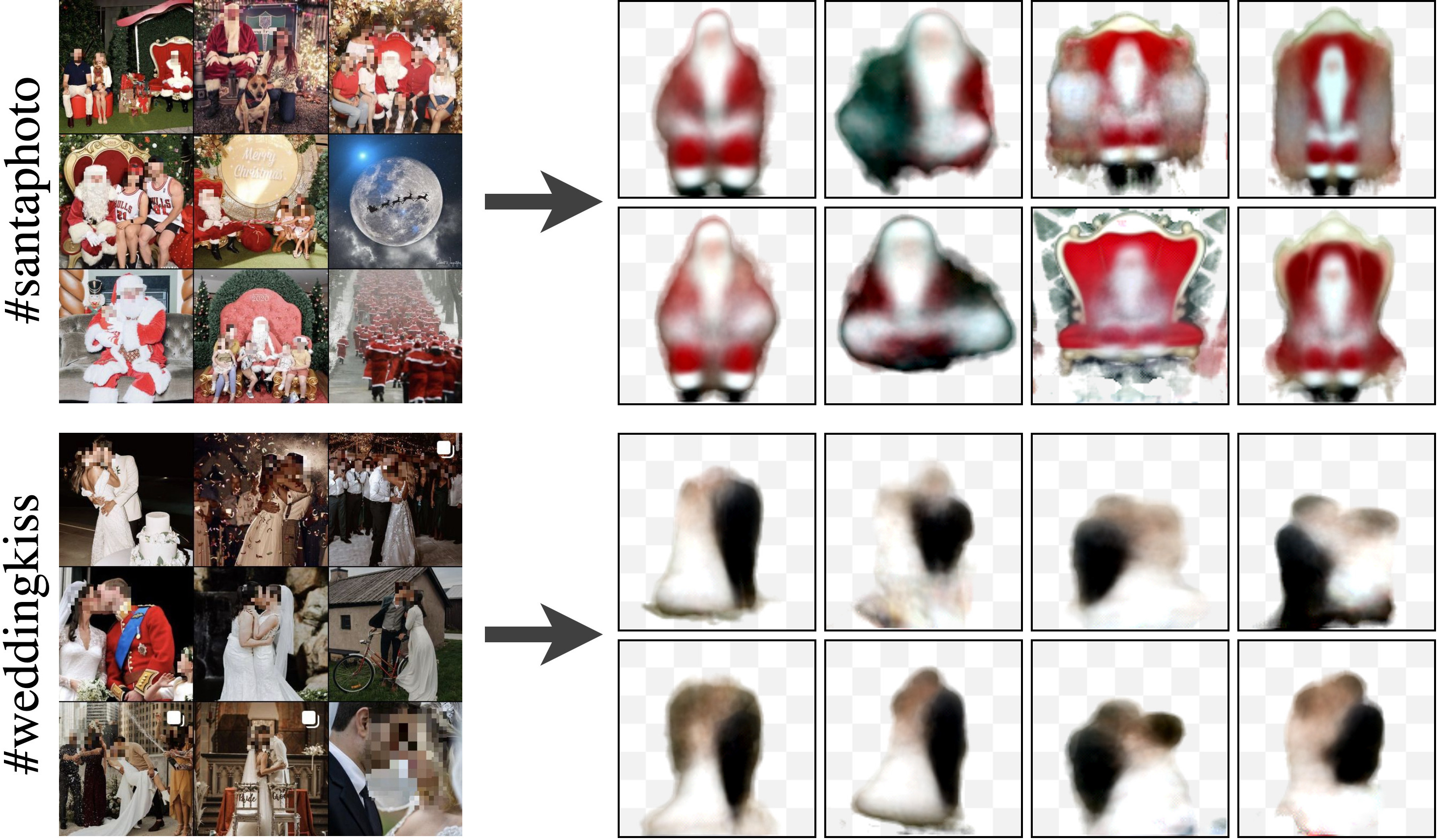}
    \vspace{-.6em}
  \end{subfigure}
  \begin{subfigure}{\linewidth}
      \centering
      \includegraphics[width=\linewidth]{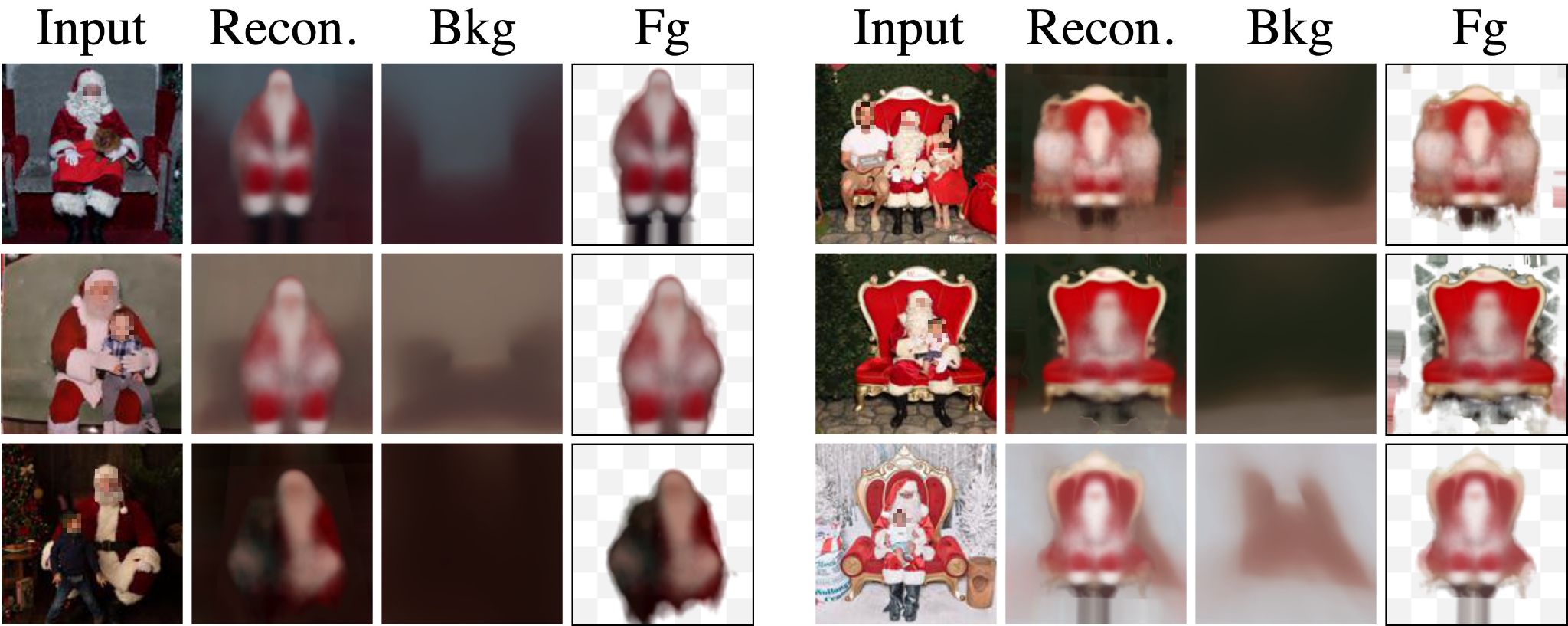}
  \end{subfigure}
  \caption{\textbf{Web image results.} We show the 8 best qualitative sprites among 40 
  discovered from Instagram collections (top) as well as decomposition results for samples 
represented by one of the sprites shown for \#santaphoto (bottom).}
  \label{fig:insta}
  \vspace{-.7em}
\end{figure}

We demonstrate our approach's robustness by visualizing sprites discovered from web image 
collections. We use the same Instagram collections introduced 
in~\cite{monnierDeepTransformationInvariantClustering2020}, where each collection is 
associated to a specific hashtag and contains around 15k images resized and center cropped to 
$128\times128$. We apply our model with 40 sprites and a background.

Figure~\ref{fig:insta} shows the 8 best qualitative sprites discovered from Instagram 
collections associated to \#santaphoto and \#weddingkiss. Even in this case where
images are mostly noise, our approach manages to discover meaningful sprites and 
segmentations with clear visual variations. For example, we can distinguish standing santas 
from seating ones, as well as the ones alone or surrounded by children. We additionally show
examples of reconstructions and image compositions for some of the 8 sprites shown for 
\#santaphoto. 

\section{Conclusion}

We have introduced a new unsupervised model which jointly learns sprites, transformations and 
occlusions to decompose images into object layers. Beyond standard multi-object synthetic 
benchmarks, we have demonstrated that our model leads to actual improvements for real image 
clustering with a 5\% boost over the state of the art on SVHN and can provide good 
segmentation results. We even show it is robust enough to provide meaningful results on 
unfiltered web image collections. Although our object modeling involves unique prototypical 
images and small sets of transformations limiting their instances diversity, we argue that 
accounting for such diversity while maintaining a category-based decomposition model is 
extremely challenging, and our approach is the first to explore this direction as far as we 
know.


\section*{Acknowledgements}
We thank Fran\c{c}ois Darmon, Hugo Germain and David Picard for valuable feedback. This work 
was supported in part by: the French government under management of Agence Nationale de la 
Recherche as part of the project EnHerit ANR-17-CE23-0008 and the "Investissements d'avenir" 
program (ANR-19-P3IA-0001, PRAIRIE 3IA Institute); project Rapid Tabasco; gifts from Adobe; 
the Louis Vuitton/ENS Chair in Artificial Intelligence; the Inria/NYU collaboration; HPC 
resources from GENCI-IDRIS (2020-AD011011697).

{\small
\bibliographystyle{ieee_fullname}
\bibliography{references}
}

\clearpage
\appendix

\twocolumn[{\Large\bf\centering Supplementary Material for\\Unsupervised Layered Image 
Decomposition into Object Prototypes\\[1.5em]}]

In this supplementary document, we provide quantitative semantic segmentation results 
(Section~\ref{sec:app_sem}), analyses of the model (Section~\ref{sec:app_anal}), training 
details (Section~\ref{sec:app_details}) and additional qualitative results 
(Section~\ref{sec:app_decompo}).

\section{Quantitative semantic segmentation results}\label{sec:app_sem}

We provide a quantitative evaluation of our approach in an unsupervised semantic segmentation 
setting.  We do not compare to state-of-the-art approaches as none of them
explicitly model nor output categories for objects. We argue that modeling categories for 
discovered objects is crucial to analyse and understand scenes, and thus advocate such 
quantitative semantic evaluation to assess the quality of any object-based image 
decomposition algorithm.

\vspace{-1em}
\paragraph{Evaluation.} Motivated by standard practices from supervised semantic segmentation 
and clustering benchmarks, we evaluate our unsupervised object semantic segmentation results 
by computing the mean accuracy (mACC) and the mean intersection-over-union (mIoU) across all 
classes (including background). We first compute the global confusion matrix on the same 320 
images used for object instance segmentation evaluation. Then, we reorder the matrix with a 
cluster-to-class mapping computed using the Hungarian 
algorithm~\cite{kuhnHungarianMethodAssignment1955}. Finally, we average accuracy and 
intersection-over-union over all classes, including background, respectively yielding mACC 
and mIoU.

\vspace{-1em}
\paragraph{Results.} Our performances averaged over 5 runs are reported in 
Table~\ref{tab:multiobj_sem}.  Similar to our results for object instance segmentation, we 
filter an outlier run out of 5 for Multi-dSprites based on its high reconstruction loss 
compared to other runs ($1.93 \times 10^{-3}$ against $\{1.51, 1.49, 1.52, 1.57\} \times 
10^{-3}$). For Tetrominoes and Multi-dSprites, our method obtains strong results thus 
emphasizing that our 2D prototype-based modeling is well suited for such 2D scene benchmarks.  
On the contrary for CLEVR6, there is still room for improvements. Although we can distinguish 
the 6 different categories from discovered sprites, such performances suggest that our model 
struggles to accurately transform the sprites to match the target object instances. This is 
expected since we do not explicitly account for neither 3D, lighting nor material effects in 
our modeling.

\begin{table}[h]
  \renewcommand{\arraystretch}{1}
  \addtolength{\tabcolsep}{-2pt}
  \caption{\textbf{Multi-object semantic discovery.} We report our mACC and mIoU performances 
  averaged over 5 runs with stddev. $K$ refers to the number of classes (including 
background) and we mark results ($\vartriangle$) where one outlier run was automatically 
filtered out.}
  \vspace{-.5em}
  \centering
  \begin{tabular}{@{}lccc@{}} \toprule
  Dataset & $K$ & mACC & mIoU\\
  \midrule
  Tetrominoes~\cite{greffMultiObjectRepresentationLearning2019} & 20 & 99.5 $\pm$ 0.2 & 99.1 
  $\pm$ 0.4\\
  Multi-dSprites~\cite{multiobjectdatasets19} & 4 & 91.3$^\vartriangle \pm$ 0.9& 
  84.0$^\vartriangle \pm$ 1.4\\
  CLEVR6~\cite{johnsonCLEVRDiagnosticDataset2017, greffMultiObjectRepresentationLearning2019} 
  & 7 & 73.9 $\pm$ 2.1 & 56.3 $\pm$ 2.9\\
  \bottomrule
  \end{tabular}
  \label{tab:multiobj_sem}
\end{table}

\section{Model analysis}\label{sec:app_anal}

\subsection{Effect of $K$} 
\begin{figure}
  \centering
  \includegraphics[width=\columnwidth]{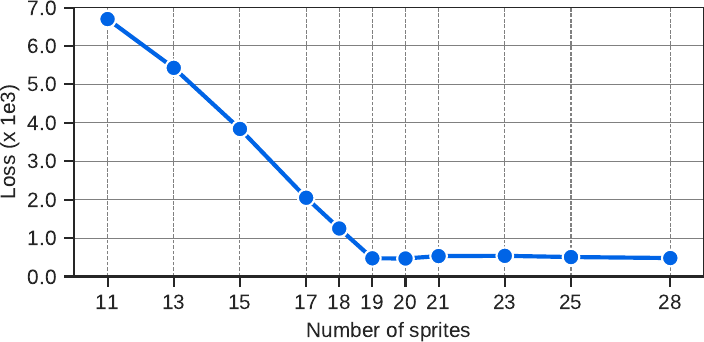}
  \caption{\textbf{Effect of $K$.} We report the loss obtained for varying number of sprites 
  on Tetrominoes, where the ground-truth number of different shapes is 19.}
  \label{fig:loss_wrt_k}
  \vspace{-.5em}
\end{figure}

Similar to standard clustering methods, our results are sensitive 
to the assumed number of sprites $K$. A purely quantitative analysis could be applied to 
select $K$, \eg in Figure~\ref{fig:loss_wrt_k} we plot the loss as a function of the number 
of sprites for Tetrominoes and it is clear an elbow method can be applied to correctly select 
19 sprites.  Qualitatively, using more sprites than the ground truth number typically yields 
duplicated sprites which we think is not that harmful. For example, we use an arbitrary 
number of sprites (40) for the Instagram collections and we have not found the discovered 
sprites to be very sensitive to this choice.

\subsection{Effect of $\lambda$} 
\begin{figure}
  \centering
  \includegraphics[width=0.8\columnwidth]{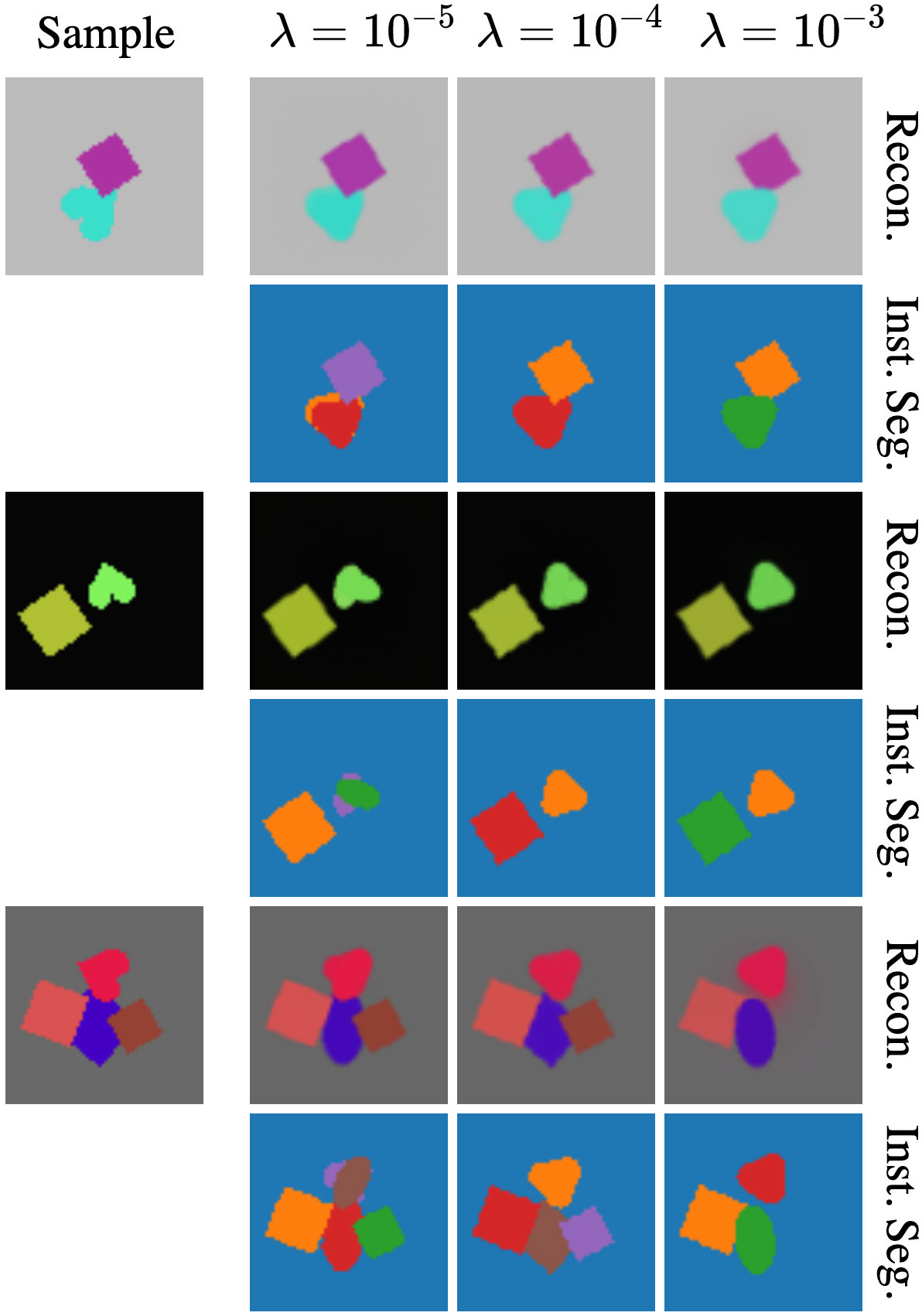}
  \caption{\textbf{Effect of $\lambda$.} We show reconstructions and instance segmentations 
  for different values of $\lambda$ on Multi-dSprites.}
  \label{fig:lambda_effect}
  \vspace{-.5em}
\end{figure}

The hyperparameter $\lambda$ controls the weight of the 
regularization term that counts the number of non-empty sprites used. In 
Figure~\ref{fig:lambda_effect}, we show qualitative results obtained for different values of 
$\lambda$ on Multi-dSprites. When $\lambda$ is zero or small (here $\lambda = 10^{-5}$), the 
optimization typically falls into bad local minima where multiple layers attempt to 
reconstruct the same object. Increasing the penalization ($\lambda = 10^{-4}$) prevents this 
phenomenon by encouraging reconstructions using the minimal number of non-empty sprites.  
When $\lambda = 10^{-3}$, the penalization is too strong and some objects are typically 
missed (last example).

\subsection{Computational cost} Training our method on Tetrominoes, Multi-dSprites and CLEVR6 
respectively takes approximately 5 hours, 3 days and 3.5 days on a single Nvidia GeForce RTX 
2080 Ti GPU.\ Our approach is quite memory efficient and for example on CLEVR6, we can use a 
batch size of up to 128 on a single V100 GPU with 16GB of RAM as opposed to 4 
in~\cite{greffMultiObjectRepresentationLearning2019} and 64 
in~\cite{locatelloObjectCentricLearningSlot2020}.

\section{Training details}\label{sec:app_details}

The full implementation of our approach and all datasets used are available at 
\href{https://github.com/monniert/dti-sprites}{https://github.com/monniert/dti-sprites}.

\subsection{Architecture}

We use the same parameter predictor network architecture for all the experiments. It is 
composed of a shared ResNet~\cite{heDeepResidualLearning2016} backbone truncated after the 
average pooling and followed by separate Multi-Layer Perceptrons (MLPs) heads predicting 
sprite transformation parameters for each layer as well as the occlusion matrix. For the 
ResNet backbone, we use mini 
ResNet-32\footnote{\href{https://github.com/akamaster/pytorch_resnet_cifar10}
{https://github.com/akamaster/pytorch\_resnet\_cifar10}} (64 features) for images smaller 
than $65\times65$ and ResNet-18 (512 features) otherwise. When modeling large numbers of 
objects ($>3$), we increase the representation size by replacing the global average pooling 
by adaptive ones yielding $4\times4\times64$ features for mini ResNet-32 and 
$2\times2\times512$ for ResNet-18. Each MLP has the same architecture, with two hidden layers 
of 128 units.

\begin{table}[t]
\renewcommand{\arraystretch}{1}
  \addtolength{\tabcolsep}{-2pt}
  \caption{\textbf{Transformation sequences used.}}
  \vspace{-.5em}
  \centering
  \small
  \begin{tabular}{@{}lccc@{}} \toprule
    Dataset & $\cT^{\,\textrm{lay}}$ & $\cT^{\,\textrm{spr}}$ & $\cT^{\,\textrm{bkg}}$\\
  \midrule
  Tetrominoes~\cite{greffMultiObjectRepresentationLearning2019} & col-pos & - & -\\
  Multi-dSprites~\cite{multiobjectdatasets19} & col-pos & sim & col\\
  CLEVR6~\cite{johnsonCLEVRDiagnosticDataset2017, greffMultiObjectRepresentationLearning2019} 
  & col-pos & proj & col\\
  \midrule
  GTSRB-8~\cite{stallkampManVsComputer2012}& - & col-proj-tps & col-proj-tps\\
  SVHN~\cite{netzer2011reading} & - & col-proj-tps & col-proj-tps\\
  Weizmann Horse~\cite{borensteinLearningSegment2004}& - & col-proj-tps & col-proj-tps\\
  Instagram collections~\cite{monnierDeepTransformationInvariantClustering2020} & - & 
  col-proj & col-proj\\
  \bottomrule
  \end{tabular}
  \label{tab:tsf}
  \vspace{-.5em}
\end{table}

\subsection{Transformation sequences}

Similar to DTI-Clustering~\cite{monnierDeepTransformationInvariantClustering2020}, we model 
complex image transformations as a sequence of transformation modules which are successively 
applied to the sprites. Most of the transformation modules we used are introduced 
in~\cite{monnierDeepTransformationInvariantClustering2020}, namely affine, projective and TPS 
modules modeling spatial transformations and a color transformation module. We augment the 
collection of modules with two additional spatial transformations implemented with spatial 
transformers~\cite{jaderbergSpatialTransformerNetworks2015}:
\begin{itemize}
  \item a \textit{positioning module} parametrized by a translation vector and a scale
    value (3 parameters) and used to model coarse layer-wise object positioning,
  \item a \textit{similarity module} parametrized by a translation vector, a scale value and 
    a rotation angle (4 parameters).
\end{itemize}

The transformation sequences used for each dataset are given in Table~\ref{tab:tsf}. All 
transformations for the multi-object benchmarks are selected to mimic the way images were 
synthetically generated. For real images, we use the col-proj-tps default sequence when the 
ground truth number of object categories is well defined and the col-proj sequence otherwise.  
Visualizing sprites and transformations helps understanding the results and adapting the 
transformations accordingly.

\subsection{Implementation details} Both sprite parameters and predictors are learned jointly 
and end-to-end using Adam optimizer~\cite{kingmaAdamMethodStochastic2015} with a 10$^{-6}$ 
weight decay on the network parameters. Background, sprite appearances and masks are 
respectively initialized with averaged images, constant value and Gaussian weights. To 
constrain sprite parameters in values close to $[0, 1]$, we use a $\softclip$ function 
implemented as a piecewise linear function yielding identity inside $[0, 1]$ and an affine 
function with $0.01$ slope outside $[0, 1]$.  We experimentally found it tends to work better 
than (i) a traditional $\clip$ function which blocks gradients and (ii) a sigmoid function 
which leads to very small gradients. Similar 
to~\cite{monnierDeepTransformationInvariantClustering2020}, we adopt a curriculum learning 
strategy of the transformations by sequentially adding transformation modules during training 
at a constant learning rate until convergence, then use a multi-step policy by multiplying 
the learning rate by $0.1$ once convergence has been reached. For the experiments with a 
single object on top of a background, we use an initial learning rate of $10^{-3}$ and reduce 
it once. For the multi-object experiments, because spatial transformations are much stronger, 
we use an initial value of $10^{-4}$ and first train global layer-wise transformations, using 
frozen sprites during the first epochs (initialized with constant value for appearances and 
Gaussian weights for masks). Once such transformations are learned, we learn sprite-specific 
transformations if any and reduce after convergence the learning rate for the network 
parameters only.  Additionally, in a fashion similar
to~\cite{monnierDeepTransformationInvariantClustering2020}, we perform sprite and predictor 
reassignment when corresponding sprite has been used for reconstruction less than $20/K \%$ 
of the images layers. We use a batch size of 32 for all experiments, except for GTSRB-8 and 
SVHN where a batch size of 128 is used. 

\subsection{Learning binary masks}
\begin{figure}[t]
  \center
  \includegraphics[width=\linewidth]{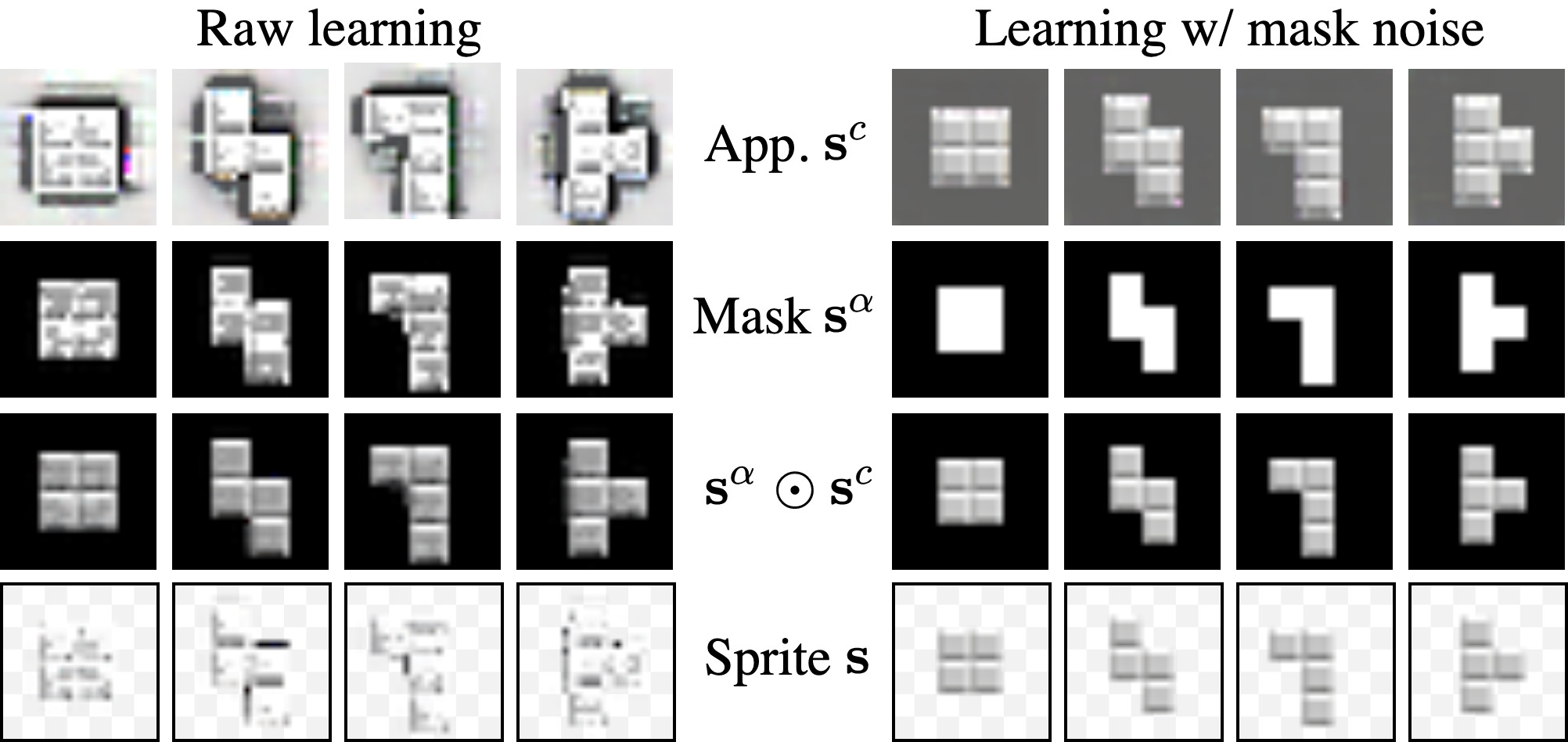}
  \caption{\textbf{Learning binary masks.} We compare results on Tetrominoes obtained with 
  (right) and without (left) injecting noise in masks.}
  \label{fig:ablation}
  \vspace{-.8em}
\end{figure}

There is an ambiguity between learned mask and color values in our raw image formation model.  
In Figure~\ref{fig:ablation}, we show examples of sprites learned following two settings: (i) 
a raw learning and (ii) a learning where we constrain mask values to be binary.
Although learned appearance images $\papp$ (first row) and masks $\mask$ (second row) are 
completely different, applying the masks onto appearances (third row) yields similar images, 
and thus similar reconstructions of sample images. However, resulting sprites (last row) 
demonstrate that the spatial extent of objects is not well defined when learning without any 
constraint.  

Since constraining the masks to binary values actually resolves ambiguity and forces clear 
layer separations, we follow the strategy adopted by 
Tieleman~\cite{tielemanOptimizingNeuralNetworks2014} and 
SCAE~\cite{kosiorekStackedCapsuleAutoencoders2019} to learn binary values, and propose to 
inject during training uniform noise $\in [-0.4, 0.4]$ into the masks before applying our 
$\softclip$.  Intuitively, such stochasticity prevents the masks from learning appearance 
aspects and can only be reduced with values close to $0$ and $1$. We experimentally found 
this approach tends to work better than (i) explicit regularization penalizing values outside 
of $\{0, 1\}$ \eg with a $x \rightarrow x^2 (1 - x)^2$ function and (ii) a varying 
temperature parameter in a sigmoid function as advocated by Concrete/Gumbel-Softmax 
distributions~\cite{maddisonConcreteDistributionContinuous2017, 
jangCategoricalReparameterizationGumbelSoftmax2017}. 

We compare our results obtained with and without injecting noise into the masks on 
Tetrominoes, where shapes have clear appearances. Quantitatively, while our full model 
reaches almost a perfect score for both ARI-FG and ARI metrics (resp. 99.6\% and 99.8\%), 
these performances averaged over 5 runs are respectively 77.8\% and 89.1\% when noise is not 
injected into the masks during learning. We show qualitative comparisons in 
Figure~\ref{fig:ablation}. Note that the masks learned with noise injection are binary and 
sharp, whereas the ones learned without noise contain some appearance patterns.

\section{Additional qualitative results}\label{sec:app_decompo}

\begin{figure}[h]
  \center
  \begin{subfigure}{\linewidth}
    \centering
    \includegraphics[width=\linewidth]{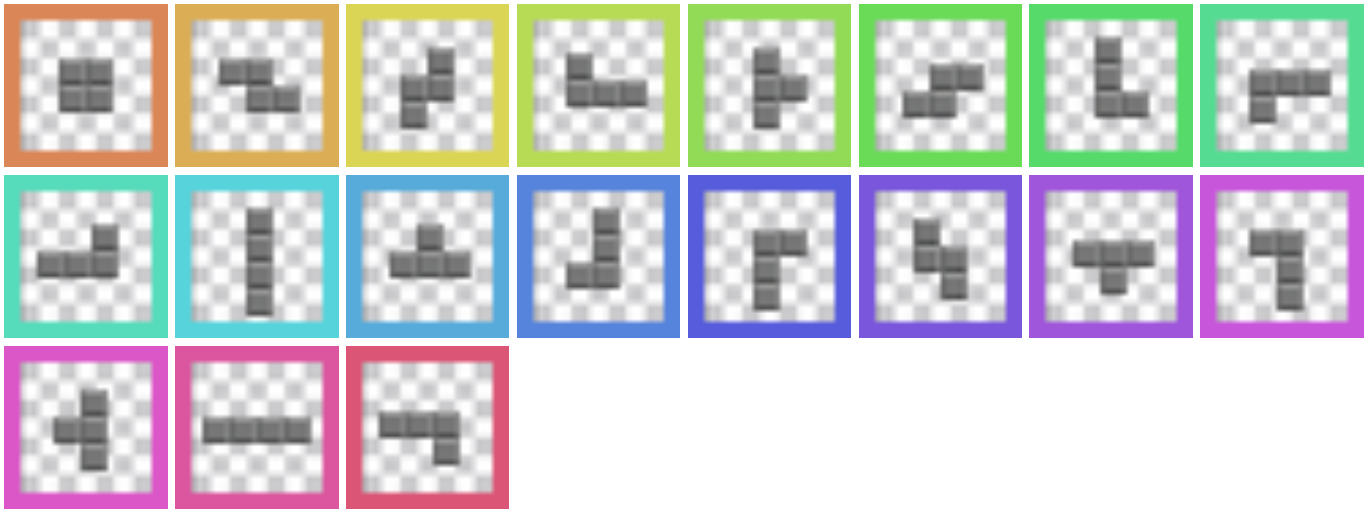}
  \end{subfigure}
  \begin{subfigure}{\linewidth}
    \vspace{0.7em}
    \includegraphics[width=\linewidth]{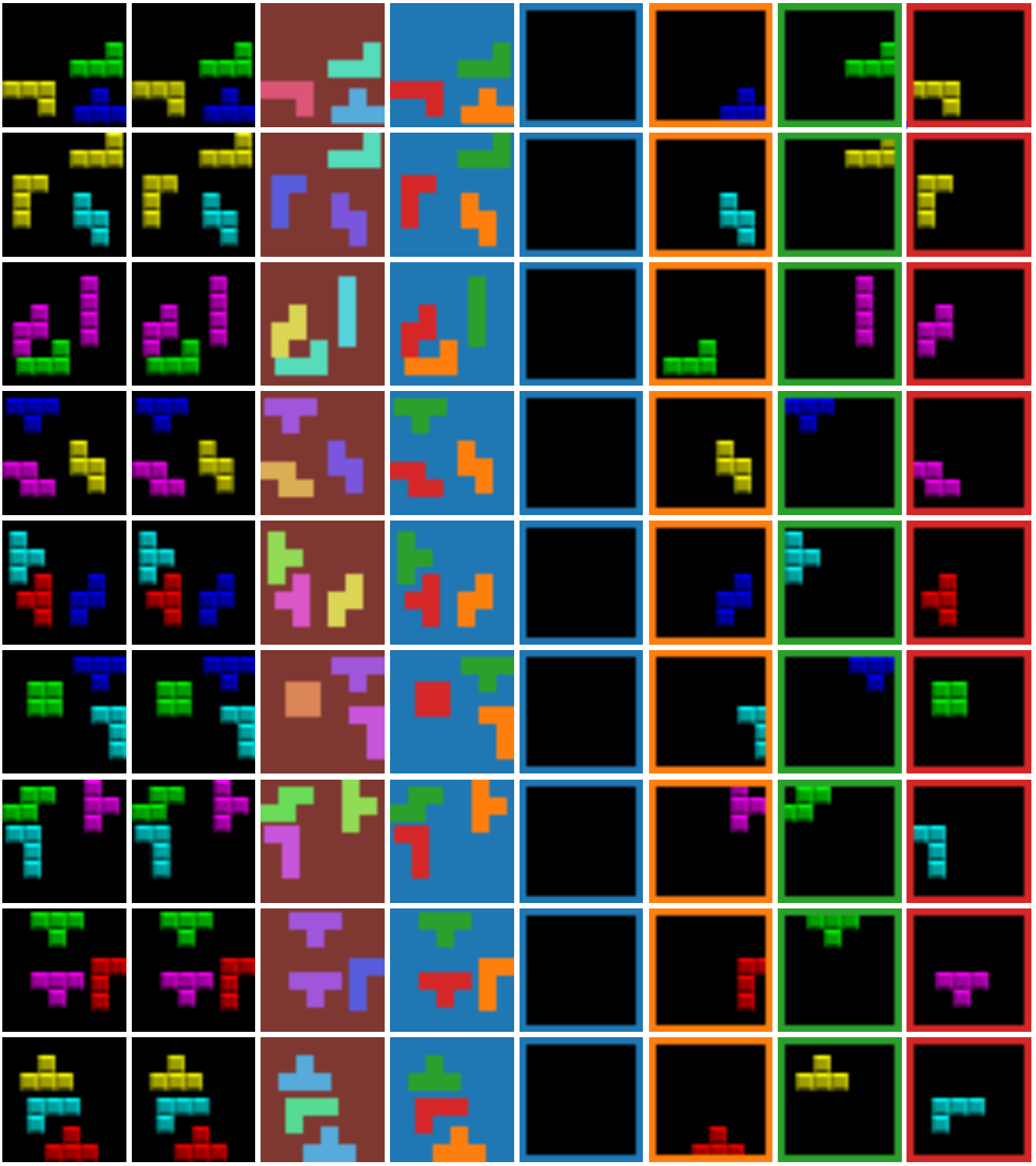}
  \end{subfigure}
  \caption{\textbf{Tetrominoes results.}  We show discovered sprites (top) and 10 random 
  decomposition results (bottom).}
  \label{fig:decompo_tetro}
\end{figure}

\begin{figure}[t]
  \center
  \begin{subfigure}{\linewidth}
    \centering
    \includegraphics[width=0.5\linewidth]{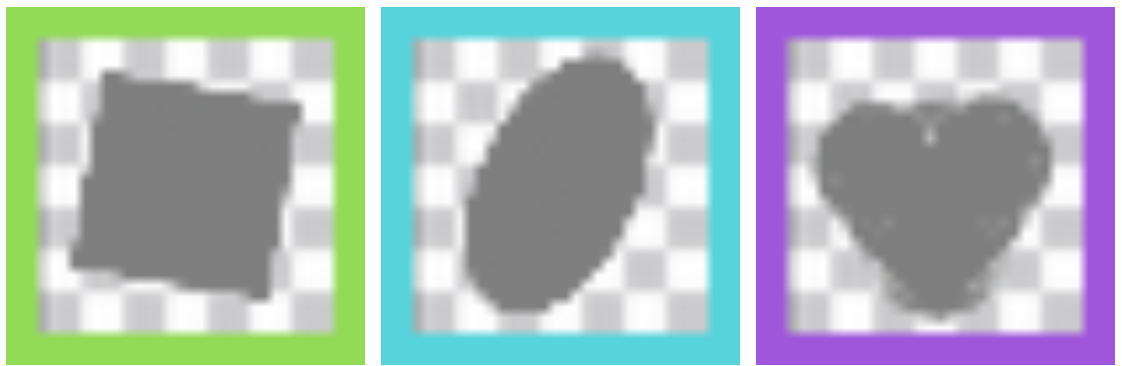}
  \end{subfigure}
  \begin{subfigure}{\linewidth}
    \vspace{0.5em}
    \includegraphics[width=\linewidth]{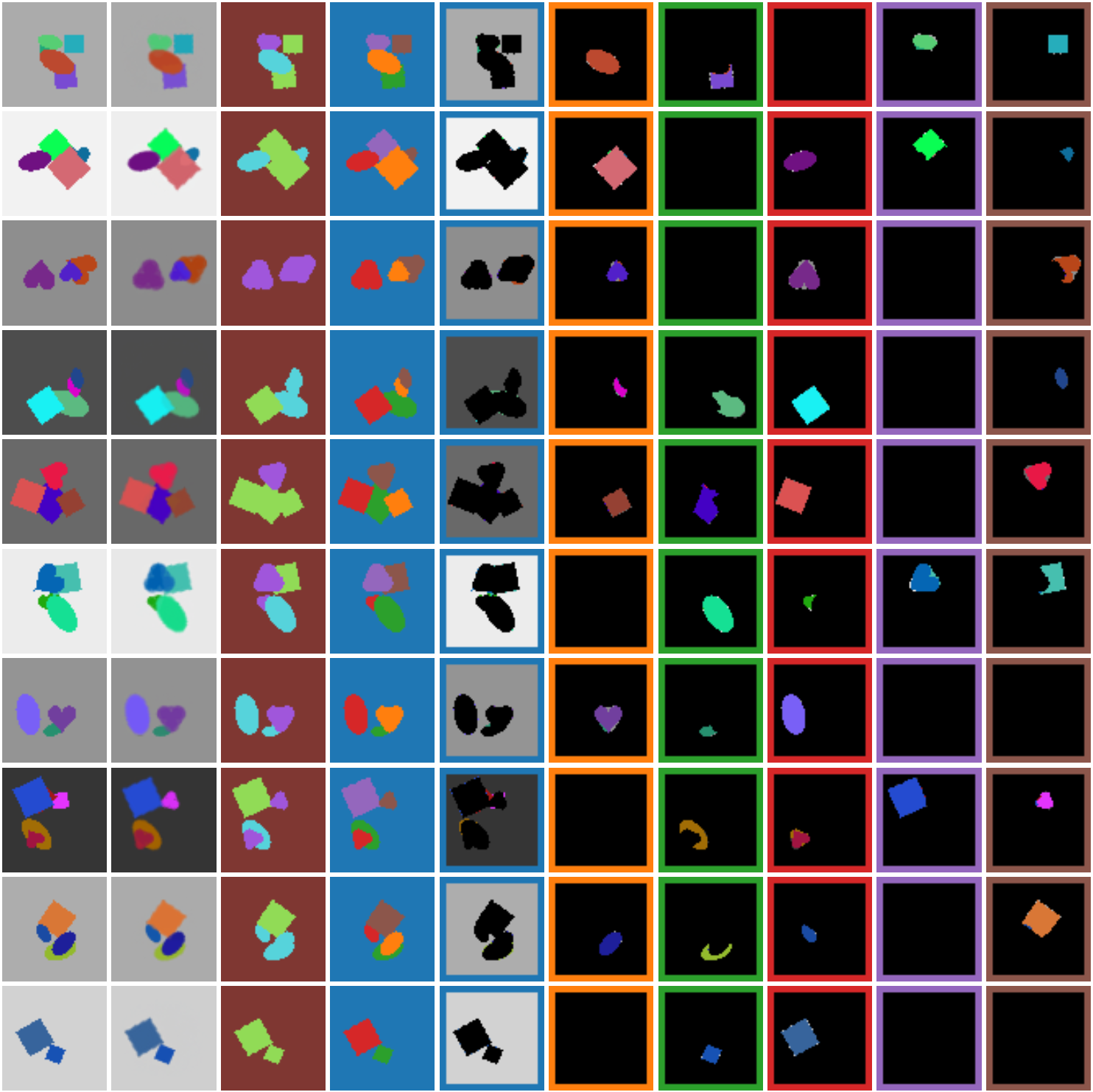}
  \end{subfigure}
  \caption{\textbf{Multi-dSprites results.} We show discovered sprites (top) and 10 random 
  decomposition results (bottom).}
  \vspace{-1.2em}
  \label{fig:decompo_multi}
\end{figure}

\begin{figure}[!b]
  \center
  \begin{subfigure}{\linewidth}
    \centering
    \includegraphics[width=\linewidth]{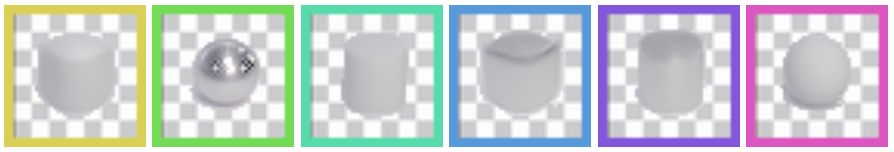}
  \end{subfigure}
  \begin{subfigure}{\linewidth}
    \vspace{0.5em}
    \includegraphics[width=\linewidth]{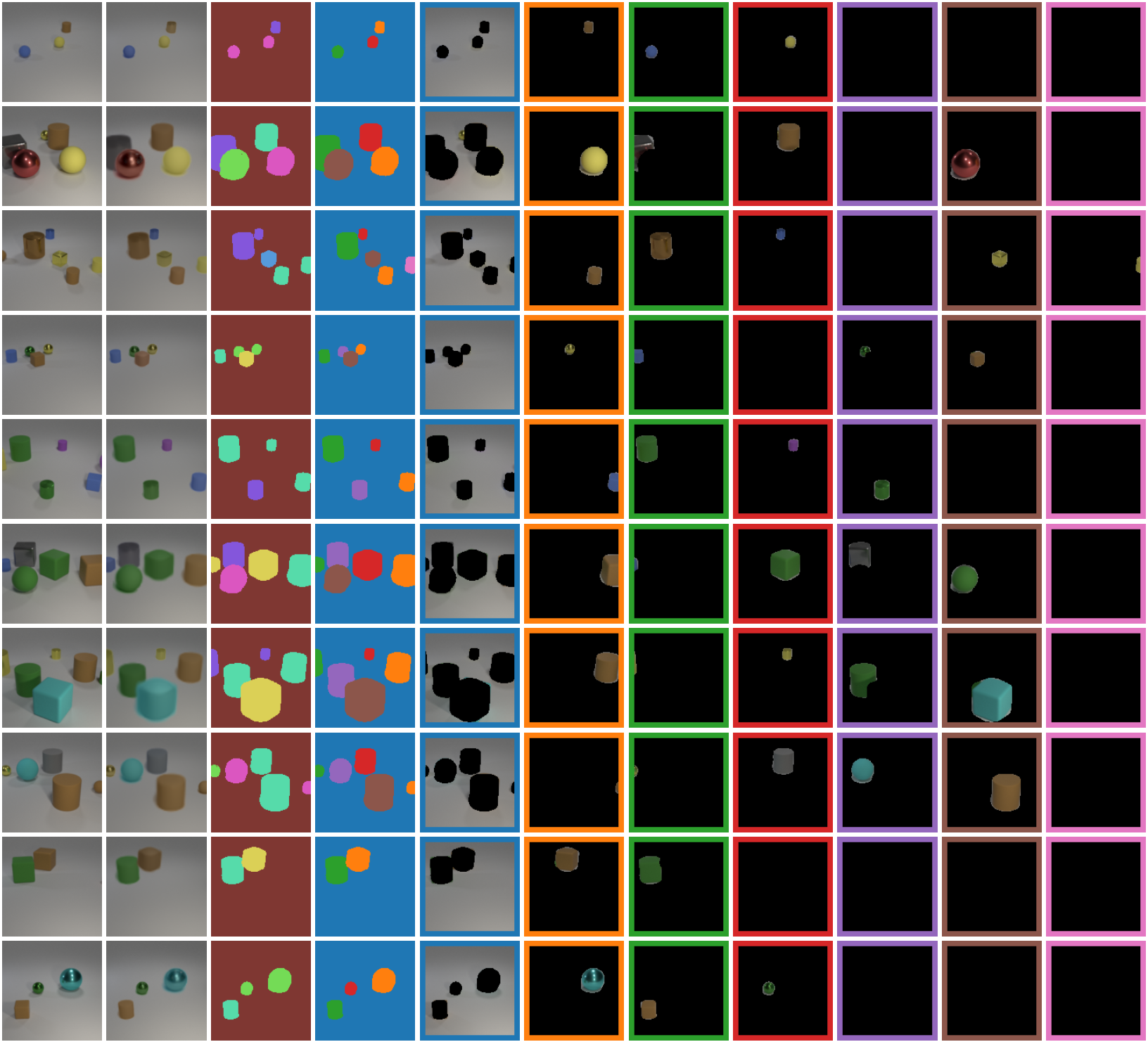}
  \end{subfigure}
  \caption{\textbf{CLEVR6 results.} We show discovered sprites (top) and 10 random 
  decomposition results (bottom).}
  \label{fig:decompo_clevr}
  \vspace{-.3em}
\end{figure}

We provide more qualitative results on the multi-object synthetic benchmarks, namely 
Tetrominoes (Fig.~\ref{fig:decompo_tetro}), Multi-dSprites (Fig.~\ref{fig:decompo_multi}) and 
CLEVR6 (Fig.~\ref{fig:decompo_clevr}). For each dataset, we first show the discovered sprites 
(at the top), with colored borders to identify them in the semantic segmentation results. We 
then show 10 random qualitative decompositions. From left to right, each row corresponds to: 
input sample, reconstruction, semantic segmentation where colors refer to the sprite border 
colors, instance segmentation where colors correspond to different object instances, and full 
image decomposition layers where the borders are colored with respect to their instance mask 
color.  Note how we manage to successfully separate the object instances as well as identify 
their categories and spatial extents.

For additional decompositions, we urge the readers to visit 
\href{https://imagine.enpc.fr/~monniert/DTI-Sprites/extra_results/index.html}
{imagine.enpc.fr/\httilde monniert/DTI-Sprites/extra\_results}.

\end{document}